%% file: example_paper.tex
\documentclass{article}

\usepackage{graphicx}
\usepackage{subfigure}
\usepackage{booktabs} % for professional tables

\usepackage{hyperref}

\usepackage[accepted]{icml2024}

\usepackage{amsmath}
\usepackage{amssymb}
\usepackage{mathtools}
\usepackage{amsthm}

\theoremstyle{plain}
\newtheorem{theorem}{Theorem}[section]

\newtheorem{lemma}[theorem]{Lemma}

\theoremstyle{definition}

\newtheorem{assumption}[theorem]{Assumption}
\theoremstyle{remark}

\usepackage[textsize=tiny]{todonotes}

\usepackage{setspace}
\usepackage{tcolorbox}
\usepackage{algorithm}
\usepackage{multirow}
\usepackage{wrapfig}
\usepackage{xcolor}
\def\code#1{\texttt{#1}}

\icmltitlerunning{Machine Vision Therapy}

\begin{document}

\twocolumn[
\icmltitle{Machine Vision Therapy: Multimodal Large Language Models Can Enhance Visual Robustness via Denoising In-Context Learning}



\icmlsetsymbol{equal}{*}
\icmlsetsymbol{thanks}{*}

\begin{icmlauthorlist}
\icmlauthor{Zhuo Huang}{thanks,usyd}
\icmlauthor{Chang Liu}{sjtu}
\icmlauthor{Yinpeng Dong}{thu}
\icmlauthor{Hang Su}{thu}
\icmlauthor{Shibao Zheng}{sjtu}
\icmlauthor{Tongliang Liu}{usyd}
\end{icmlauthorlist}

\icmlaffiliation{usyd}{Sydney AI Centre, The University of Sydney, Sydney, Australia}
\icmlaffiliation{sjtu}{Institute of Image Communication and Network Engineering, Shanghai JiaoTong University, Shanghai, China}
\icmlaffiliation{thu}{Dept. of Comp. Sci. and Tech., Institute for AI, Tsinghua-Bosch Joint ML Center, THBI Lab, BNRist Center, Tsinghua University, Beijing, 100084, China}

\icmlcorrespondingauthor{Tongliang Liu}{tongliang.liu@sydney.edu.au}

\icmlkeywords{OOD Robustness, Multimodal Large Language Model}

\vskip 0.3in
]



\printAffiliationsAndNotice{\icmlThanksContribution}  % leave blank if no need to mention equal contribution

\input{sec/0_abstract}    
\input{sec/1_introduction}

\input{sec/2_related_work}
\input{sec/3_methodology}

\input{sec/4_experiments}
\input{sec/5_conclusion}

\bibliography{example_paper}
\bibliographystyle{icml2024}

%%%%%%%%%%%%%%%%%%%%%%%%%%%%%%%%%%%%%%%%%%%%%%%%%%%%%%%%%%%%%%%%%%%%%%%%%%%%%%%
%%%%%%%%%%%%%%%%%%%%%%%%%%%%%%%%%%%%%%%%%%%%%%%%%%%%%%%%%%%%%%%%%%%%%%%%%%%%%%%
% APPENDIX
%%%%%%%%%%%%%%%%%%%%%%%%%%%%%%%%%%%%%%%%%%%%%%%%%%%%%%%%%%%%%%%%%%%%%%%%%%%%%%%
%%%%%%%%%%%%%%%%%%%%%%%%%%%%%%%%%%%%%%%%%%%%%%%%%%%%%%%%%%%%%%%%%%%%%%%%%%%%%%%
\clearpage
\appendix

\input{sec/X_suppl}
%%%%%%%%%%%%%%%%%%%%%%%%%%%%%%%%%%%%%%%%%%%%%%%%%%%%%%%%%%%%%%%%%%%%%%%%%%%%%%%
%%%%%%%%%%%%%%%%%%%%%%%%%%%%%%%%%%%%%%%%%%%%%%%%%%%%%%%%%%%%%%%%%%%%%%%%%%%%%%%

\end{document}

%% file: sec/0_abstract.tex
\vspace{-1mm}
\begin{abstract}
Although pre-trained models such as Contrastive Language-Image Pre-Training (CLIP) show impressive generalization results, their robustness is still limited under Out-of-Distribution (OOD) scenarios. Instead of undesirably leveraging human annotation as commonly done, it is possible to leverage the visual understanding power of Multi-modal Large Language Models (MLLMs). However, MLLMs struggle with vision problems due to task incompatibility, thus hindering their effectiveness. In this paper, we propose to effectively leverage MLLMs via Machine Vision Therapy which aims to rectify erroneous predictions of specific vision models. By supervising vision models using MLLM predictions, visual robustness can be boosted in a nearly unsupervised manner. Moreover, we propose a Denoising In-Context Learning (DICL) strategy to solve the incompatibility issue. Concretely, by examining the noise probability of each example through a transition matrix, we construct an instruction containing a correct exemplar and a probable erroneous one, which enables MLLMs to detect and rectify the incorrect predictions of vision models. Under mild assumptions, we theoretically show that our DICL method is guaranteed to find the ground truth. Through extensive experiments on various OOD datasets, our method demonstrates powerful capabilities for enhancing visual robustness under many OOD scenarios.
\vspace{-2mm}
\end{abstract}

%% file: sec/1_introduction.tex
\section{Introduction}
\label{sec:introduction}
\vspace{-2mm}
Pre-trained vision models such as Vision Transformers (ViT)~\cite{dosovitskiy2020image, liu2021swin, wang2021pyramid} with Contrastive Language-Image Pretraining (CLIP)~\cite{chen2023clip2scene, radford2021learning, li2021supervision, li2022blip, wang2024clips, zheng2023enhancing} have been widely used thanks to their strong generalization performance meanwhile effectively avoiding training vision models from scratch. But when deployed to Out-of-Distribution (OOD) scenarios~\cite{dong2023benchmarking, kong2023robo3d, hendrycks2016baseline, hong2024improving, li2024dynamic, peng2023sam, wang2019learning, zhu2023understanding}, their recognition performance could be seriously degraded~\cite{shu2023clipood}. Downstream fine-tuning has been a common practice to regain the generalizability~\cite{goyal2023finetune, wortsman2022robust}, but it requires additional label acquisition through human labor, which is undesirable for large-scale applications. 

\begin{figure}
\vspace*{-0mm}
\includegraphics[width=\linewidth]{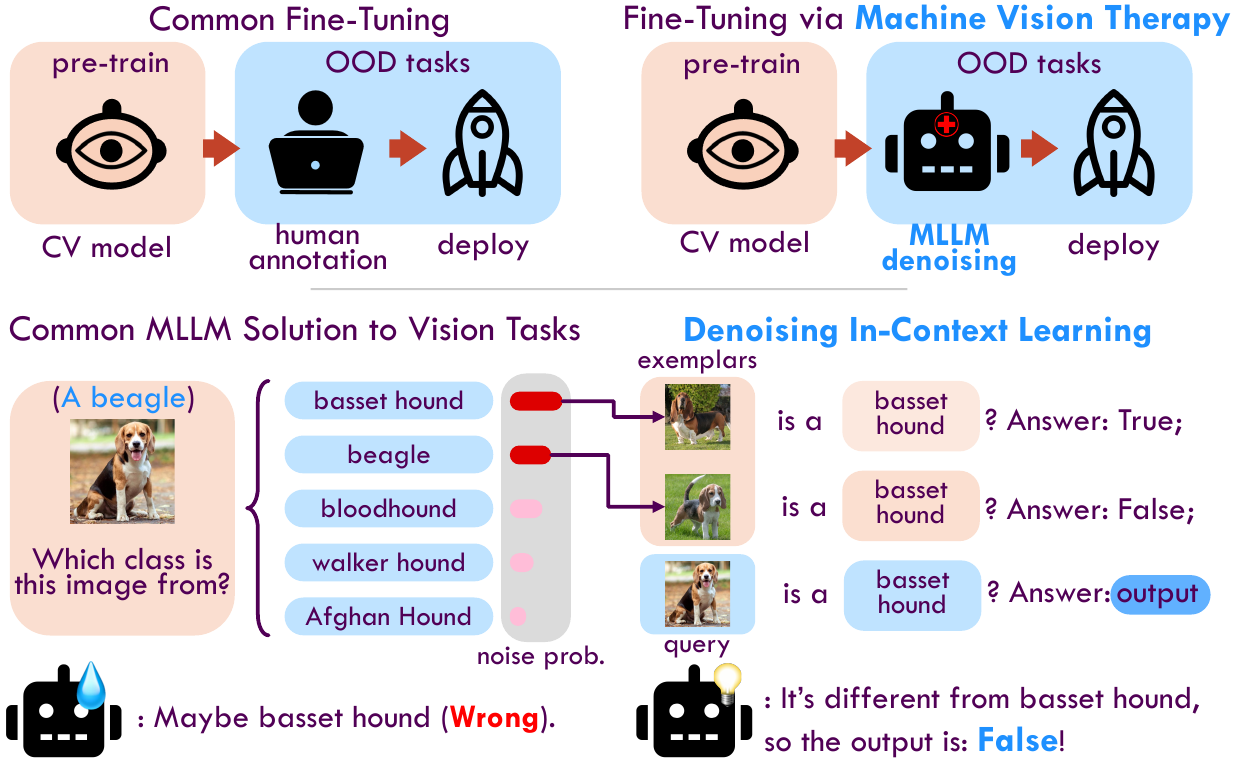}
\vspace{-8mm}
\caption{\small Illustration of our methodology: Upper row: Comparison between common fine-tuning process and fine-tuning via Machine Vision Therapy. Our method potentially eliminates the necessity for human-annotation by leveraging the knowledge from MLLMs. Lower row: Comparison between previous MLLM solution to vision tasks and Denoising In-Context Learning strategy. Instead of considering all classes, our method make predictions by presenting a pair of positive and negative exemplars.}
\label{fig:motivation}
\vspace*{-6mm}
\end{figure}

Fortunately, the thriving Multi-modal Large Language Models (MLLMs)~\cite{alayrac2022flamingo, awadalla2023openflamingo, chen2024towards, gong2023multimodal, li2023blip, li2023otter, liu2023llava, ye2023mplug, zhu2023minigpt}, which take advantage of the few-shot learning ability of Large Language Models (LLM)~\cite{brown2020language, chung2022scaling, floridi2020gpt, openai2023gpt4, scao2022bloom, touvron2023llama, touvron2023llama2, zheng2023judging}, have manifested powerful capabilities on understanding visual information with language interpretations, and excelled at recognizing novel objects in multimodal tasks such as image captioning, visual question answering, visual reasoning, etc. Considering the vulnerability of vision models under OOD situations, here we hope to refine vision models by leveraging the knowledge of MLLMs, as shown in the upper row of Figure~\ref{fig:motivation}. However, due to the difficulty of aligning the text generation process with visual recognition tasks\footnote{In this paper, we mainly focus on classification task.}~\cite{alayrac2022flamingo, wang2022git}, MLLMs struggle with generating correct answers that match the ground-truth class names, thus underperforming the current dominant contrastive paradigms, even when employing them as own vision encoders~\cite{alayrac2022flamingo, awadalla2023openflamingo, huang2023language, wang2022git, zhai2023investigating}.

Focusing on enhancing the robustness of vision models, in this paper, we propose to effectively leverage MLLMs to conduct \textbf{Machine Vision Therapy (MVT)} which aims to diagnose and rectify the error predictions through a novel Denoising In-Context Learning (DICL) strategy. Then, we utilize the rectified supervision to guide the fine-tuning process in downstream OOD problems. Specifically, rather than giving a set of options to ask MLLMs for the exact answer~\cite{alayrac2022flamingo, huang2023language, zhai2023investigating}, we show that it is sufficient to query for the ground truth by using only two exemplars, \textit{i.e.}, 1) a correct one that demonstrates the exact match between a query class name with its image example and 2) an erroneous one that combines the same query class with an image from the most confusing category for the vision model. Since the erroneous predictions are essentially label noise, hence we draw inspiration from learning with noisy labels~\cite{han2018co, liu2015classification, lin2022we, lin2023cs, natarajan2013learning, wu2024mitigating, wu2023making, xia2020part, xia2020robust, yao2020dual, yao2021instance, yao2023better, yuan2024early}. Particularly, we can find the erroneous categories by estimating a transition matrix that captures the probability of one class being mistaken as another. By feeding the two exemplars, MLLMs can be instructed to leverage their few-shot learning power to distinguish the semantically similar images that are easily misclassified by vision models, as shown in the lower row of Figure~\ref{fig:motivation}. To process such instructions, we leverage the multi-modal in-context learning ability of several existing MLLMs~\cite{chen2023lightweight, li2023mimicit, yasunaga2023retrieval, zhao2023mmicl} to realize our methodology. After the error predictions are diagnosed and rectified, vision models can be further fine-tuned to enhance their OOD robustness on downstream data distribution. Through a comprehensive empirical study on many challenging datasets and their OOD variants, such as ImageNet~\cite{deng2009imagenet}, WILDS~\cite{koh2021wilds}, and DomainBed~\cite{gulrajani2021search}, we carefully validate the effectiveness of our method and demonstrate its superiority under various OOD scenarios on many well-known vision models.

\vspace{-2mm}
To sum up, our contributions are three-fold:
\vspace{-1mm}
\begin{itemize}
	\vspace{-2.5mm}
	\item We design a novel Machine Vision Therapy paradigm to enhance computer vision models by effectively leveraging the knowledge of MLLMs without needing additional label information.
	
	\vspace{-2.5mm}
	\item We propose a Denoising In-Context Learning strategy to successfully align MLLMs with vision tasks.
	
	\vspace{-2.5mm}
	\item Through comprehensive quantitative and qualitative studies on many well-known datasets, we demonstrate that the proposed method can enhance:
    \textcolor[rgb]{0.05, 0.09, 0.58}{1)} generalization on both ID and OOD data,
    \textcolor[rgb]{0.05, 0.09, 0.58}{2)} robustness against domain shift,
    \textcolor[rgb]{0.05, 0.09, 0.58}{3)} robustness against common corruptions,
    \textcolor[rgb]{0.05, 0.09, 0.58}{4)} performance on recognizing fine-Grained attributes,
    \textcolor[rgb]{0.05, 0.09, 0.58}{5)} robustness against spurious correlations,
    \textcolor[rgb]{0.05, 0.09, 0.58}{6)} detection on prediction errors and OOD data.
\end{itemize}

%% file: sec/3_methodology.tex
\vspace{-5mm}
\section{Methodology}
\label{sec:method}
\vspace{-2mm}

\begin{figure*}[t]
\vspace*{-1mm}
\centering
\includegraphics[width=0.93\linewidth]{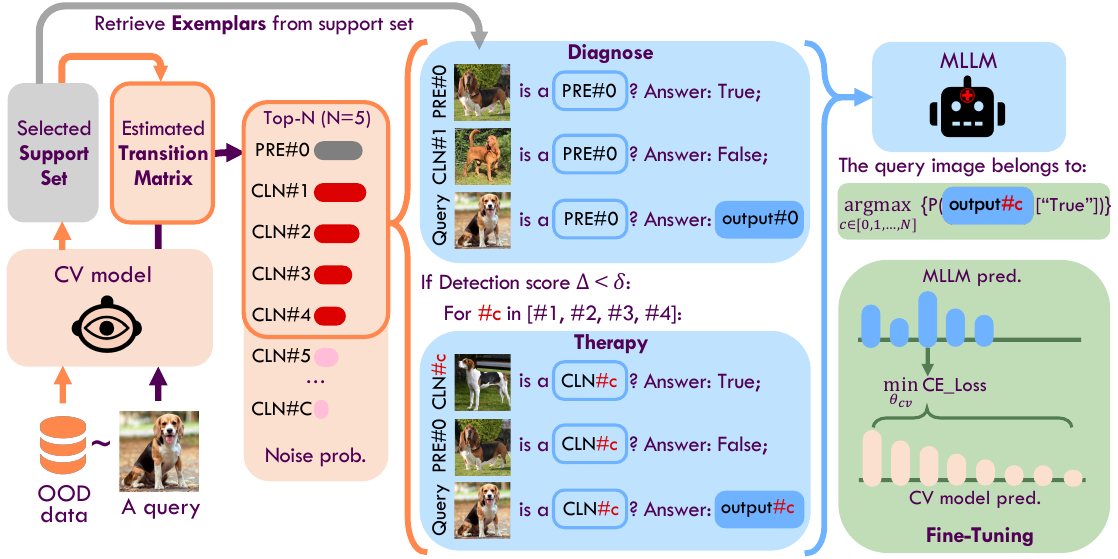}
\vspace*{-5mm}
\caption{\small Workflow of our Machine Vision Therapy: The orange part demonstrates the Transition Matrix Estimation, the blue part indicates the Denoising In-Context Learning process, and the green part illustrates the Fine-Tuning of vision models.}
\label{fig:framework}
\vspace*{-3mm}
\end{figure*}

In this section, we carefully demonstrate the Machine Vision Therapy process which mainly contains three components, namely Transition Matrix Estimation, Denoising In-Context Learning, and Fine-Tuning of vision models. Next, we demonstrate problem setting and framework overview.

\vspace{-1mm}
\subsection{Problem Formulation and Overview}
\vspace{-1mm}
\label{sec:formulation_overview}
Generalizing to Out-of-Distribution tasks has been a challenging topic in computer vision problems, where we normally have a vision model parameterized by $\theta_{cv}\in\Theta_{cv}$ pre-trained on massive labeled in-distribution (ID) data $\mathcal{D}^{id}=\{x^{id}_i, y^{id}_i\}_{i=0}^{m}\in\mathcal{X}\times\mathcal{Y}$, where $\mathcal{Y}=\mathbb{R}^C$. Here each ID example is sampled from a joint distribution, \textit{i.e.}, $(X^{id}, Y^{id})\sim p^{id}$, where $X^{id}$ and $Y^{id}$ stand for variables. After pretraining, we can assume the conditional distribution $P(Y^{id}|X^{id})$ can be perfectly captured by the inference function $\tilde{y}^{id}=f_{\theta_{cv}}(x^{id})$, where $\tilde{y}^{id}$ is the prediction. In OOD tasks, we are given a set of unlabeled examples $\mathcal{D}^{ood}=\{x^{ood}_i\}_{i=0}^{n}$ whose element $x^{ood}\in\mathcal{X}$ is drawn from an unknown data distribution $p^{ood}$. Due to the change of downstream task, some factors that affect the data generating process are shifted, causing a difference between $p^{ood}$ and $p^{id}$, further hindering the label prediction, \textit{i.e.}, $\tilde{y}^{ood}=f_{\theta_{cv}}(x^{ood})\not\sim P(Y^{ood}|X^{ood})$, where $Y^{ood}$ is the unknown ground truth. Fortunately, having been observed with extraordinary low-shot generalization capability, we leverage MLLM with parameters $\theta_{mllm}\in\Theta_{mllm}$ to enhance the OOD robustness of vision models.

Our framework is illustrated in Figure~\ref{fig:framework} and our problem can be formulated as follows:
\vspace{-2.2mm}
\begin{align}
    \min_{\theta_{cv}}\mathcal{L}(f_{\theta_{cv}}, z);\ z&=\left[\theta_{mllm}((X_c^{+}\!, Y_c^{+}); (X_c^{-}\!, Y_c^{+}); \!X_i)\right]_{c}^N\!; \nonumber \\
    Y_c&=T[c; \arg\!\max [f_{\theta_{cv}}(X_i)]],
\vspace{-3mm}
\label{eq:problem_formulation}
\end{align}
where $X_i^{+}$ and $X_i^{-}$ denotes the positive and negative exemplars, respectively, $X_i$ is the query image, and $T$ is the transition matrix. Intuitively, when a distribution shift occurs, the emerging prediction errors are essentially label noises that can be captured by estimating a transition matrix. Hence, by focusing on calibrating the examples with high noise probabilities, the visual robustness of downstream tasks can be improved effectively. In particular, we feed all OOD data into the vision model to obtain the noisy prediction distribution $P(\tilde{Y}^{ood}|X^{ood})$, based on which we can effectively estimate $T$ and provide exemplars to instruct MLLM\footnote{Although some manual annotation is required, we show in later experiments that our strategy has an acceptable labeling workload and demonstrates superior performance to vanilla fine-tuning on the support set. Furthermore, the support set is \textbf{ not used for parameter tuning} in our method, so our fine-tuning does not actually use any human annotation for training.}. Further, we conduct machine vision therapy to find the possible ground truth for $X_i$ based on the MLLM output $z$. Finally, $z$ is leveraged to minimize $\mathcal{L}$ to optimize $\theta_{cv}$. Next, we explain the details of each process.

% based on the MLLM output, we first fine-tune $\theta_{cv}$ by minimizing a loss function $\mathcal{L}$. Specifically,

% Then, based on the estimated $T$, we use the label prediction ``PRE\#0'' of $X_i$ to obtain top-$N$ noisy class list, in which we fix the first element to be ``PRE\#0'' and sort the rest elements ``CLN\#c'' based on noise probability. 

\vspace{-1mm}
\subsection{Transition Matrix Estimation}
\vspace{-1mm}
\label{sec:t_estimation}
The distribution shift from OOD data $x^{ood}$ leads to unreliable label prediction $\tilde{y}^{ood}$, which is highly unreliable due to instance-dependent feature noises~\cite{li2024instant, xia2020part} as shown in Section~\ref{sec:ablation}. Hence, in order to capture the relationship between $\tilde{Y}^{ood}$ and $Y^{ood}$, we leverage a transition matrix $T\in [0, 1]^{C\times C}$~\cite{liu2015classification, natarajan2013learning, xia2019anchor} which satisfies $P(Y^{ood}|X^{ood})=T^\top P(\tilde{Y}^{ood}|X^{ood})$. However, estimating such a transition matrix is difficult without access to any noisy label supervision or strong assumption~\cite{liu2015classification, xia2019anchor}. Therefore, we propose a simple yet effective sample selection approach to construct a support set with clean labels. Specifically, we rank all OOD data within each class based on their prediction confidence, \textit{i.e.}, $\max_{c}\left[f_{\theta_{cv}}(x^{ood})\right]_c$, where $\left[\cdot\right]_c$ denotes the value of the $c$-th entry. From the sorted dataset $\{x_1^{ood, c}, x_2^{ood, c}, \cdots, x_{\frac{n}{C}}^{ood, c}\}_{c=1}^{C}$, we uniformly sample $\rho$ examples per class, where $\rho$ is the labeling budget, \textit{i.e.}, $\mathcal{D}^{supp}=\{\{x_{j\times\frac{n}{\rho C}}^{ood, c}\}_{j=1}^{\rho}\}_{c=1}^{C}$. In this way, we can effectively model the noisy posterior $P(\tilde{Y}^{ood}|X^{ood})$. Then, through an acceptable labeling process\footnote{We experimentally show that when there is a distribution shift between $\mathcal{D}^{supp}$ and $\mathcal{D}^{ood}$, the proposed method can still perform effectively. As a result, it is unnecessary to conduct the labeling process on each practical task. Instead, we can just use the existing support set to instruct most of OOD tasks.}, we can obtain the clean label posterior $P(Y^{ood}|X^{ood})$, thus effectively estimating the transition matrix $T$. Finally, the noise transition probability $T\left[:;\arg\max[f_{\theta_{cv}}]\right]$ of a query image can be obtained by indexing $T$ through its current prediction.

\vspace{-1mm}
\subsection{Denoising In-Context Learning}
\vspace{-1mm}
\label{sec:dicl}
Thanks to the previously obtained noise probability list $T\left[:;\arg\max[f_{\theta_{cv}}]\right]$, we can further decide which one is the possible ground truth through DICL. In particular, we only consider the classes of the top-$N$ noise probability as potential candidates. If the label prediction denoted by ``PRE\#0'' is not in the candidates, we would fix it in the first place. Further, we conduct \textit{Diagnosing} which decides the fidelity of the current prediction, and \textit{Therapy} which finds the possible ground truth.

\vspace*{-1mm}
\paragraph{Diagnosing.} Since the inference time of MLLMs is non-trivial, it is necessary to avoid redundant analysis on confident examples. Hence, to examine the fidelity of vision model predictions, our Diagnosing focuses on answering whether a query image belonging to class ``PRE\#0'' is ``True''. Specifically, we retrieve from $\mathcal{D}^{supp}$ to obtain one exemplar image belonging to ``PRE\#0'', and another exemplar image belonging to the class with the largest noise transition probability ``CLN\#1''\footnote{The performance of retrieve strategy is carefully studied in Section~\ref{sec:performance_analysis}.}. Then, combined with the query image $X_q$, an in-context instruction is constructed:
\vspace*{-2mm}
\begin{tcolorbox}
Question: This image $<$IMG\_PRE\#0$>$ shows a photo of $<$PRE\#0$>$, True or False? Answer: True;\\
\vspace{-2.5mm}

Question: This image $<$IMG\_CLN\#1$>$ shows a photo of $<$PRE\#0$>$, True or False? Answer: False;\\
\vspace{-2.5mm}

Question: This image $<$IMG\_Query$>$ shows a photo of $<$PRE\#0$>$, True or False? Answer:
\end{tcolorbox}
The symbols $<$IMG\_PRE\#0$>$, $<$IMG\_CLN\#1$>$, and $<$IMG\_Query$>$ are replace tokens for the image features of exemplars from ``PRE\#0'' and ``CLN\#1'', and $X_q$, respectively. The first exemplar acts as the positive one to show MLLMs the true image from class ``PRE\#0'', and the second exemplar shows the negative one to show the highly probable false image from ``CLN\#1''. Then, based on the $X_q$ and ``PRE\#0'', MLLMs can effectively judge the correctness by outputting $z_0$:
\vspace{-1mm}
\begin{equation}
	\!\!\!z_0\!=\!\theta_{mllm}((X_{PRE\#0},\! Y_{PRE\#0}); (X_{CLN\#1},\! Y_{PRE\#0});\! X_q)\!.\!\!
	\label{eq:diagnose}
\vspace{-1mm}
\end{equation}
To enable further quantitative analysis, we obtain the logits of ``True'' and ``False'' tokens from the MLLM output $z_0$ followed by a softmax function, \textit{i.e.}, $z_0:=\text{softmax}(\left[z_0[\text{True}], z_0[\text{False}]\right])$. Finally, we combine $z_0[\text{True}]$ and the prediction confidence of the vision model to obtain a detection score $\Delta$:
\vspace{-1mm}
\begin{equation}
	\Delta = \frac{1}{2}(z_0[\text{True}]+\max_{c}\left[f_{\theta_{cv}}\right]_c(x^{ood})).
	\label{eq:detection_score}
\vspace{-1mm}
\end{equation}
If $\Delta$ is larger than a threshold $\delta$, we assume the current prediction ``PRE\#0'' is correct\footnote{Detailed analysis is shown in Section~\ref{sec:performance_analysis}.}, otherwise, we conduct the next Therapy process.

\vspace{-2mm}
\paragraph{Therapy.} 
During therapy, we continue to use the instruction template above and traverse across the rest clean class candidates. Particularly, for each iteration $c$ in $N-1$ trials, we choose ``CLN\#c'' as the positive class and ``PRE\#0'' as the negative class, whose exemplars are correspondingly retrieved from $\mathcal{D}^{supp}$ to construct the prompt. Then, it is fed into MLLM to output whether the query image belongs to the class ``CLN\#c'', \textit{i.e.}, $z_c=\theta_{mllm}((X_{CLN\#c}, Y_{CLN\#c}); (X_{PRE\#0},\! Y_{CLN\#c});\! X_q)$,\!\! let $z_c := \text{softmax}(\left[z_c[\text{True}], z_c[\text{False}]\right])$. As a result, we can decide the final prediction through:
\vspace{-3mm}
\begin{equation}
	y_{mllm} = \arg\max \left[z_c[\text{True}]\right]_{c=0}^N.
	\label{eq:mllm_pred}
\vspace{-3mm}
\end{equation}

As shown in Section~\ref{sec:experiments}, the performance of MLLM prediction shows strong performance in many OOD scenarios. However, we still cannot directly employ MLLMs for inference, due to three main reasons: \textcolor[rgb]{0.05, 0.09, 0.58}{1)} Non-negligible inference time: Since current MLLMs cannot handle large-batch data, it would be unimaginably slower (\textit{e.g.}, 1000$\times$) when using MLLMs rather than vision models; \textcolor[rgb]{0.05, 0.09, 0.58}{2)} High requirements for computation: Inference through MLLM takes up huge memory of GPU. For MLLMs using large LLMs such as LLaMA-13B, it requires distributed inference on less advanced devices; \textcolor[rgb]{0.05, 0.09, 0.58}{3)} Model privacy issue: Many MLLMs are highly sensitive with limited accessibility, therefore. Hence, we propose to fine-tune vision models based on the prediction of MLLMs.

\subsection{Fine-Tuning of Vision Models}
\label{sec:fine-tuning}
After obtaining the MLLM prediction $y_{mllm}$, we propose to optimize vision models through the following objective:
\vspace{-3mm}
\begin{equation}
min_{\theta_{cv}}\mathcal{L}_{ce}(f_{\theta_{cv}}, y_{mllm}),
\label{eq:fine-tuning}
\vspace{-1mm}
\end{equation}
where $\mathcal{L}_{ce}(\cdot)$ denotes the cross-entropy loss. Here we summarize our methodology in Algorithm~\ref{alg:MVT}. Further, we can directly deploy the fine-tuned vision models to OOD tasks whose effectiveness is demonstrated in Section~\ref{sec:experiments}.

\renewcommand{\algorithmicrequire}{\textbf{Input:}}
\renewcommand{\algorithmicensure}{\textbf{Output:}}
\begin{algorithm}[tb]
	\scriptsize
	\caption{\small Machine Vision Therapy.}
	\label{alg:MVT}
	\begin{algorithmic}[1]
		\REQUIRE Pre-trained vision model $\theta_{cv}$, MLLM $\theta_{mllm}$, OOD dataset $\mathcal{D}^{ood}$.
		\STATE Uniformly sample $\rho C$ examples from confidence-sorted $\mathcal{D}^{ood}$ to construct support set $\mathcal{D}^{supp}$;
		\STATE Estimate transition matrix $T$;
		\COMMENT{\textit{\color{black!60} Section~\ref{sec:t_estimation}}}
		\FOR{$i \in 0,1, \cdots, n$}
		\STATE Based on the label prediction $\tilde{y}_i^{ood}$ to obtain the noisy transition probability $T\left[:;\arg\max[f_{\theta_{cv}}]\right]$;
		\STATE Conduct Diagnosing through Eq.~\ref{eq:diagnose} and compute detection score $\Delta$ through Eq.~\ref{eq:detection_score};
		\IF {$\Delta>\delta$}
		\STATE Accept current prediction;
		\ELSE 
		\STATE Conduct Therapy and obtain MLLM prediction through Eq.~\ref{eq:mllm_pred};
		\COMMENT{\textit{\color{black!60} Section~\ref{sec:dicl}}}
		\STATE Based on the MLLM prediction, conduct fine-tuning through Eq.~\ref{eq:fine-tuning}.
		\COMMENT{\textit{\color{black!60} Section~\ref{sec:fine-tuning}}}
		\ENDIF
		\ENDFOR
	\end{algorithmic}
\end{algorithm}

\subsection{Theoretical Analysis}
We denote the MLLM is pretrained over a distribution $p$ defined by a latent concept $\phi\in\Phi$. During DICL, there are $n$ examples to form a prompt $S_n$ which are sampled from a prompt distribution $p_{prompt}$ defined by concept $\phi^*\in\Phi$. To justify the proposed DICL strategy, based on the theoretical framework proposed by~\citet{xie2021explanation}, we show that when MLLM achieving the most probable $z$ based on the given prompt $S_n$ and query image-text pair $x_q$-$y$ under a concept $\phi^*$, the corresponding $y$ is the same as the one found from $p_{prompt}$, which is $y_q$ that matches with $x_q$.

\begin{assumption}[Distribution consistency]
$\forall (x_q, y_q)\sim p_{prompt}, p(x_q, y_q)=p_{prompt}(x_q, y_q)$.
\vspace{-3mm}
\end{assumption}

Moreover, the assumptions from~\citet{xie2021explanation} also hold, then we have the following Theorems:

\begin{theorem}
    Assume that the above assumptions hold, if for all $\phi\in\Phi$, $\phi\neq\phi^*$, the concept $\phi^*$ satisfies the distinguishability condition: $\sum_{j=1}^k KL_j(\phi^*\|\phi) > \epsilon_{start}^{\phi} + \epsilon_{delim}^{\phi}$, then as $n\rightarrow\infty$, the prediction according to the pretraining distribution is
    \vspace{-3mm}
    \begin{equation}
        \arg\max_y p(y\vert S_n, x_q, \phi^*) \rightarrow \arg\max_y p_{prompt}(y\vert x_q).
    \vspace{-3mm}
    \end{equation}
    Thus, the in-context predictor $f_n$ achieves the optimal $0-1$ risk: $\lim_{n\rightarrow\infty}\mathcal{L}_{0-1}(f_n)=\inf_f\mathcal{L}_{0-1}(f)$.
    \label{theo:optimal_y}
\end{theorem}

\begin{lemma}
    Under the same condiction of Theorem~\ref{theo:optimal_y}, the prediction $z$ according to the pretraining distribution is
    \vspace{-3mm}
    \begin{equation}
        \!\!\!\arg\!\max_z p(z\vert S_n,\! x_q,\! y_q,\! \phi^*) \!\!\rightarrow\! \arg\!\max_z p_{prompt}(z\vert x_q,\! y_q).\!\!\!\!
    \vspace{-5mm}
    \end{equation}
    \label{lemm:optimal_z_supp}
\end{lemma}

\begin{theorem}
    Assume that the above assumptions hold, as $n\rightarrow\infty$, when achieving the largest prediction probability of $z$ given prompt under concept $\phi^*$, the corresponding class description $y$ follows the same $y$ obtained from the prompt distribution:
    \vspace{-3mm}
    \begin{equation}
        \!\!\!\arg\!\max_y p(z\vert S_n,\! x_q,\! y,\! \phi^*)\! \rightarrow \!\arg\!\max_y p_{prompt}(z\vert x_q,\! y).\!\!\!
    \end{equation}
    \vspace{-8mm}
    \label{theo:dicl}
\end{theorem}

Please see the \textbf{appendix} for proof. We can see that if $n$ is large enough, the MLLM prediction $z$ achieves the largest value when $y_q$ is the exact match to $x_q$. As a result, we can justify that only when we feed the positive image-text pair to the MLLM, the prediction $z$ is the largest among all other combinations between $x_q$ and $y\in\mathcal{Y}, y\neq y_q$.

%% file: sec/4_experiments.tex
\vspace{-2mm}
\section{Experiments}
\vspace{-2mm}
\label{sec:experiments}
In this section, we first provide our experimental details. Then we conduct quantitative comparisons with the state-of-the-art vision models. Finally, we conduct ablation studies and analyses to qualitatively validate our method.

\begin{table*}[t]
\small
\vspace{-2mm}
\centering
\caption{\small Classification accuracy (\%) of baseline CLIP models and our method on 5 ID datasets and 5 OOD datasets. The baseline methods includes ViT-L from CLIP~\cite{radford2021learning} and ViT-g from EVA~\cite{EVA}, VQA, and Vanilla FT.}
\vspace{0mm}
\setlength{\tabcolsep}{2.3mm}
\begin{tabular}{l|l|ccccc|ccccc}
\toprule
\multirow{2}{*}{Arch}&\multirow{2}{*}{Method}&\multicolumn{5}{c|}{ID}&\multicolumn{5}{c}{OOD}\\
\cline{3-12}
&& IN-Val & IN-V2 & CIFAR10 & CIFAR100 & MNIST & IN-A & IN-R & IN-SK & IN-V & iWildCam \\
\hline\hline
RN50 & \multirow{3}{*}{CLIP} & 59.7 & 52.6 & 71.5 & 41.9 & 58.5 & 23.9 & 60.7 & 35.4 & 31.1 & 8.2 \\
RN101 & & 61.7 & 56.2 & 80.8 & 48.8 & 51.6 & 30.2 & 66.7 & 40.9 & 35.4 & 12.3 \\
ViT-B & & 62.9 & 56.1 & 89.9 & 65.0 & 47.9 & 32.2 & 67.9 & 41.9 & 30.5 & 10.9 \\
\hline
\multirow{5}{*}{ViT-L} & CLIP & 75.8 & 70.2& 95.6 & 78.2 & 76.4 & 69.3 & 86.6 & 59.4 & 51.8 & 13.4 \\
        & VQA & 64.9 & 59.9 & \underline{97.6} & \bf83.2 & 56.7 & 66.0 & 87.3 & 56.9 & 56.2 & 13.3 \\
        & Vanilla FT & \underline{76.1} & \bf70.8  & 96.1 & 80.3 & \underline{77.5} & 70.8 &  87.5 &  \underline{60.0} &  53.6 & \underline{15.2} \\
        & MVT & 75.2 & \bf70.8 & \bf97.9 & 78.9 & 53.0 & \underline{71.2} & \underline{88.1} & 59.0 & \underline{62.1} & \bf25.0 \\
        & +FT & \bf76.9 & \underline{70.5} & 96.7 & \underline{82.0} & \bf79.2 & \bf75.1 & \bf89.5 & \bf61.4 & \bf68.8 & - \\
\hline
\multirow{5}{*}{ViT-g} & EVA & 78.8 & 71.2 & 98.3 & 88.8 & 62.2 & 71.9 & 91.4 & 67.7 & 64.9 & 21.9 \\
     & VQA & 64.3 & 59.6 & 97.9 & 84.5 & 55.7 & 64.6 & 87.4 & 58.2 & 59.2 & 19.7 \\
     & Vanilla FT & 78.9 & \underline{71.8} & \underline{98.7} & \underline{89.1} & 62.9 & 72.7 & \underline{91.6} & \underline{68.1} & 65.6 & \underline{22.4} \\
     & MVT & \bf79.1 & 71.6 & 98.1 & 89.0 & \underline{63.2} & \underline{73.2} & 91.4 & 67.9 & \underline{66.3} & \bf25.1 \\
     & +FT & \underline{79.0} & \bf72.2 & \bf98.9 & \bf91.2 & \bf65.7 & \bf75.5 & \bf92.8 & \bf68.6 & \bf70.6 & - \\
     \bottomrule
\end{tabular}
\label{tab:1}
\vspace{-3mm}
\end{table*}

\vspace{-2mm}
\subsection{Experimental Setup}
\vspace{-2mm}
\paragraph{Datasets.} In our experiments, we use well-known ID datasets including ImageNet-1K~\cite{deng2009imagenet} validation dataset, ImageNet-V2~\cite{recht2019imagenet}, CIFAR10~\cite{krizhevsky2009learning}, CIFAR100~\cite{krizhevsky2009learning} and MNIST~\cite{lecun1998gradient}. We also evaluate OOD generalization on datasets that are commonly considered OOD ones, ImageNet-A~\citep{hendrycks2021natural}, ImageNet-R~\citep{hendrycks2021many}, mageNet-Sketch~\citep{wang2019learning}, ImageNet-V~\citep{dong2022viewfool}, iWildCam~\cite{wilds2021}, and DomainBed~\cite{gulrajani2020search}.

\vspace{-2mm}
\paragraph{Models and baselines.} For vision backbone, we employ CLIP models~\cite{radford2021learning} and utilize ViT-L/14 and ViT-g~\cite{zhai2022scaling} from EVA~\cite{EVA} as the vision model to be enhanced. For the MLLM backbone, we consider two existing works MMICL~\cite{zhao2023mmicl} and Otter~\cite{li2023otter} that possess multimodal ICL ability. We also assess the performance of CLIP variants on ResNet50, ResNet101, and ViT-B/32 as alternative vision encoders in the appendix. Additionally, we conduct Visual Question Answering (VQA) to directly ask MLLMs the class of query images. Moreover, we conduct vanilla fine-tuning (Vanilla FT) using only $D^{supp}$ as a baseline. The performance of using MLLM prediction is denoted as MVT, and our fine-tuning result is denoted as FT.

\begin{table*}[t]
\vspace{-2mm}
\small
\caption{\small Classification accuracy (\%) of baseline CLIP models and our method on 4 subsets of DomainBed datasets. The baseline methods includes ViT-L from CLIP~\cite{radford2021learning} and ViT-g from EVA~\cite{EVA}, VQA, and Vanilla FT.}
\vspace{0mm}
\centering
\setlength{\tabcolsep}{1.35mm}
\begin{tabular}{l|l|cccc|cccc|cccc|ccccc|c}
\toprule
& Datasets & \multicolumn{4}{c|}{VLCS} & \multicolumn{4}{c|}{PACS} & \multicolumn{4}{c|}{OfficeHome} & \multicolumn{5}{c|}{DomainNet} & \multirow{2}{*}{Avg} \\
\cline{3-19}
& method & 0 & 1 & 2 & 3 & 0 & 1 & 2 & 3 & 0 & 1 & 2 & 3 & 0 & 1 & 2 & 3 & 4 \\
\hline
\hline
\multirow{4}{*}{\rotatebox{90}{ViT-L}} & CLIP & 74.9 & 83.5 & 80.3 & 74.5 & 97.8 & 97.4 & 97.5 & \underline{99.4} & 87.7 & 92.7 & 85.7 & 85.6 & 61.1 & 62.1 & 60.2 & 78.4 & 51.1 & 80.6 \\

& Vanilla FT & 78.8 & 85.2 & 83.4 & 77.0 & \bf98.0 & 97.6 & 97.7 & 99.6 & \underline{87.9} & 93.1 & 87.1 & 86.9 & \underline{62.0} & \underline{62.5} & \underline{60.5} & 78.5 & 51.9 & 81.6 \\

& MVT & \underline{83.8} & \underline{89.0} & \underline{87.2} & \underline{80.3} & 97.6 & \underline{97.5} & \bf98.0 & \underline{99.4} & 87.7 & \underline{93.4} & \underline{89.0} & \underline{88.5} & 61.3 & 62.1 & 60.4 & \underline{78.7} & \underline{53.4} & \underline{82.8} \\

& +FT & \bf84.2 & \bf89.8 & \bf87.9 & \bf82.5 & \bf98.0 & \bf98.2 & \bf98.0 & \bf99.8 & \bf90.9 & \bf95.0 & \bf90.9 & \bf90.8 & \bf62.5 & \bf63.8 & \bf62.4 & \bf80.1 & \bf54.0 & \bf84.0 \\

\hline

\multirow{4}{*}{\rotatebox{90}{ViT-g}} & EVA & 72.5 & 80.0 & 79.8 & 72.8 & \underline{99.0} & \underline{98.8} & \underline{98.9} & \underline{99.8} & 90.5 & 94.2 & 88.6 & 88.7 & 61.4 & 64.7 & 61.2 & 81.6 & 54.9 & 81.6 \\

& Vanilla FT & 75.5 & 82.3 & 82.1 & 75.6 & 98.9 & 98.7 & \underline{98.9} & \underline{99.8} & \underline{90.6} & \underline{94.5} & 89.2 & 89.0 & 61.5 & 64.9 & 61.3 & 81.8 & 54.8 & 82.3 \\

& MVT & \underline{81.2} & \underline{86.6} & \underline{86.1} & \underline{79.5} & 98.2 & 98.0 & 98.0 & 99.4 & 89.7 & 93.8 & \underline{89.7} & \underline{89.1} & \bf62.2 & \bf65.0 & \underline{61.6} & \bf82.3 & \underline{56.1} & \underline{83.3} \\

 & +FT & \bf83.7 & \bf89.5 & \bf86.9 & \bf82.0 & \bf99.1 & \bf98.9 & \bf99.0 & \bf100.0 & \bf91.6 & \bf95.1 & \bf90.7 & \bf90.6 & \underline{61.9} & \underline{64.8} & \bf63.2 & \underline{81.9} & \bf56.6 & \bf84.4 \\
\bottomrule
\end{tabular}
\label{tab:2}
\vspace{-5mm}
\end{table*}

\vspace{-2mm}
\paragraph{Settings.}
For model evaluation, we randomly select 5000 images independently from the ImageNet validation set (IN-Val), ImageNet-V2 (IN-V2), ImageNet-A (IN-A), ImageNet-R (IN-R), ImageNet-Sketch (IN-SK), ImageNet-V (IN-V) and 10000 images independently from CIFAR10, CIFAR100, MNIST to constitute the test samples. Additionally, we select $3$ images per category to construct a support set to provide in-context exemplars. In the main paper, we evaluate iWildCam from WILDS and VLCS, PACS, OfficeHome, and DomainNet from DomainBed. For the details of implementation, we choose the top-$6$ noisy classes to conduct MVT. Concretely, we set the threshold $\delta=0.6$ to diagnose incorrect predictions, then we retrieve exemplars from the support set based on the most similar logit prediction to query images. For each round of DICL, we repeat the process for $3$ times and average the model predictions. During fine-tuning, we optimize the vision models for $3$ epochs using Adam and SGD optimizers for ViT-L and ViT-g, respectively. Other details and datasets are shown in \textbf{Appendix}.

\vspace{-2mm}
\subsection{Quantitative Comparison}
\vspace{-2mm}
First, we compare our MVT method with well-known vision models under both ID and OOD scenarios. As shown in Table~\ref{tab:1}, we can see that our method with fine-tuning denoted as ``+FT'' achieves better performance in most settings. Specifically, on ``IN-V'', our method with fine-tuning can significantly surpass both CLIP and EVA for $17\%$ and $6\%$, respectively. Moreover, on ``IN-A'', our method achieves $4.3\%$ and $2.8\%$ performance improvement over the second-best method on both ViT-L and ViT-g backbone, respectively. We can also observe that even without fine-tuning, the prediction accuracy of MLLM denoted by ``MVT'' can still surpass all baselines on most scenarios, which denotes the strong performance enhancement of our MVT fine-tuning on vision models. Note that we did not provide fine-tuning on iWildCam because most of the predictions are incorrect. Though MVT can still achieve the best result, the vision encoders could be misled by erroneous decisions during the fine-tuning process.

Furthermore, we consider domain shift by leveraging DomainBed datasets. Specifically, for each dataset, we leave one domain out as a test dataset and fine-tune on rest domains. By comparing two state-of-the-art vision backbones ViT-L and ViT-g, we show the performance comparison in Table~\ref{tab:2}. As we can see, both MVT and MVT with fine-tuning can significantly surpass the baseline methods. For some scenarios such as the PACS dataset, our method can achieve nearly $100\%$ performance. Moreover, in several scenarios in the VLCS dataset, both our MVT and fine-tuning can achieve almost $10\%$ improvements. Additionally, we find that our method with fine-tuning largely surpasses vanilla fine-tuning baseline on both Tables~\ref{tab:1} and~\ref{tab:2}. Hence, we can conclude that our learning strategy can indeed provide effective supervisions which enhances vision robustness under distribution shift.

\vspace{-2mm}
\subsection{Ablation Study}
\vspace{-2mm}
\label{sec:ablation}
In this part, we conduct ablation studies to analyze each module of MVT by using ViT-L backbone vision model.

\begin{figure}[h]
    \centering
    \includegraphics[width=\linewidth]{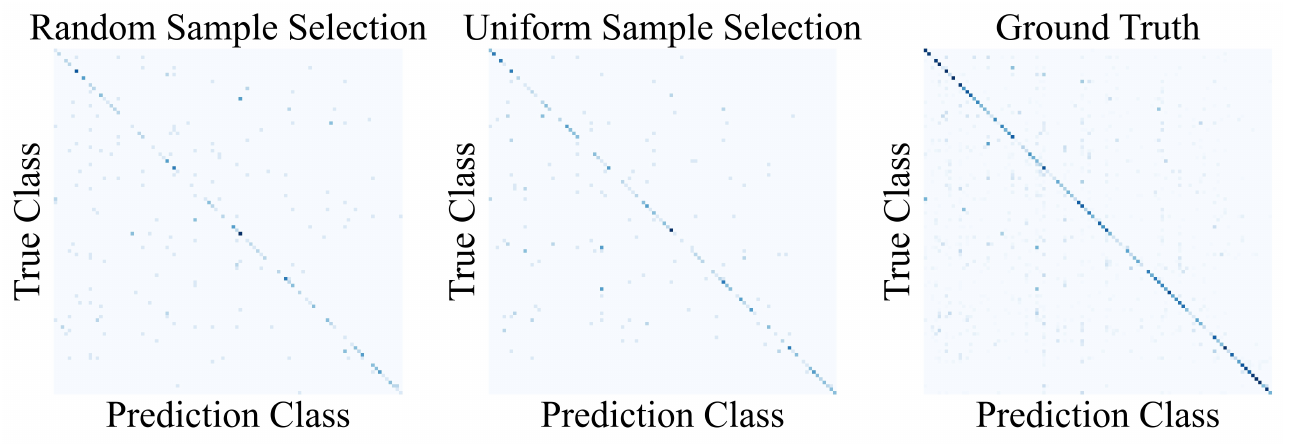}
\vspace*{-6mm}
\caption{\small Ablation study on transition matrix estimation by comparing our method with random sampling and ground truth.}
\label{fig:estimate}
\vspace*{-0mm}
\end{figure}

\vspace{-5mm}
\paragraph{Ablation Study on Transition Matrix Estimation.}
To validate the performance of transition matrix estimation, we compare our confidence-based uniform sampling strategy to a random sampling baseline. The result on the ImageNet-V dataset is shown in Figure~\ref{fig:estimate}. To quantitatively show the superiority of our method, we compute the $\ell_2$ norm of the difference between one estimation and ground truth which indicates the fidelity of the estimation. As a result, our estimation is much more accurate by achieving $3.83$ norm, compared to $4.46$ of random sampling.

\setlength{\intextsep}{1.4pt}
\setlength{\columnsep}{2pt}

\vspace{-5mm}
\paragraph{Ablation Study on Choosing Noisy Classes.}
Further, we justify the choice of using a transition matrix to obtain the noisy classes. As a comparison, we use the top-$6$ predictions as the therapy candidates and show the results in Table~\ref{tab:4}. We can see on all datasets, our method can outperform the opponent with non-trivial improvements. Therefore, leveraging the transition matrix to find the potential noisy classes is more effective than using prediction.

\begin{table}[h]
\small
\vspace{0mm}
\centering
\caption{\small Performance comparison between choosing noisy classes through transition matrix (MVT) and using Top-$N$ predictions.}
\vspace{-0mm}
\setlength{\tabcolsep}{1.5mm}
\begin{tabular}{lcccccc}
\toprule
& IN-A & IN-SK & IN-Val & IN-R & IN-V2 & IN-V \\
\hline
\hline
Top-$N$ Pred. & 60.3 & 58.4 & 74.2 & 85.3 & 67.7 & 58.3 \\
MVT & 65.5 & 59.0 & 75.1 & 86.0 & 70.7 & 61.6 \\
\bottomrule
\label{tab:4}
\end{tabular}
\vspace{-3mm}
\end{table}

\begin{figure}[t]
\vspace{-2mm}
\centering
\includegraphics[width=\linewidth]{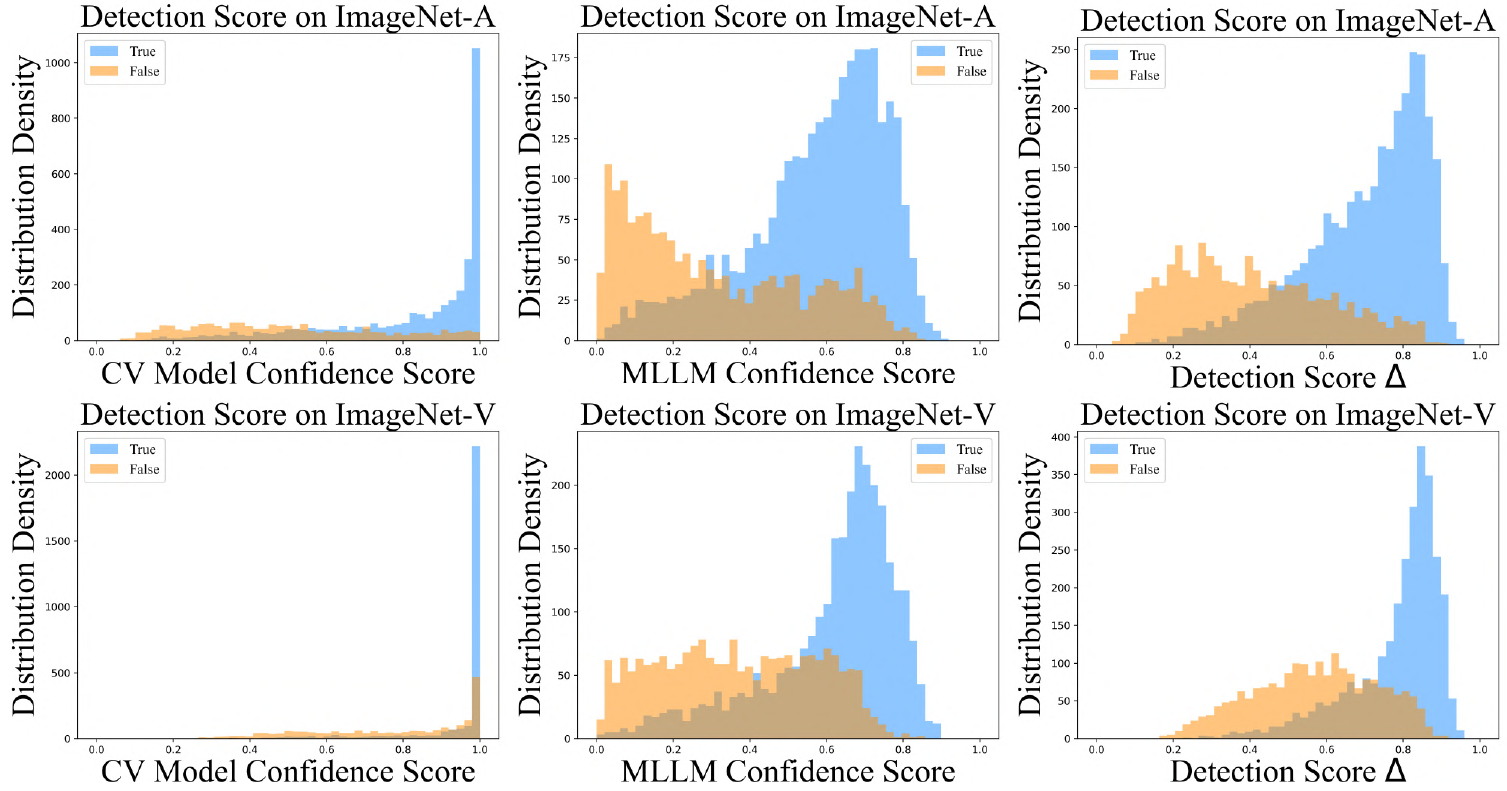}
\vspace*{-8mm}
\caption{\small Ablation study on detection score distribution. Upper: ImageNet-A; Lower: ImageNet-V.}
\label{fig:detection_score}
\vspace*{-3mm}
\end{figure}

\vspace{-0mm}
\paragraph{Ablation Study on Detection Score.}
To analyze the proposed detection score on conducting diagnosing, we show the distribution of prediction confidence provided by the vision model, MLLM, and our detection score $\Delta$ in Figure~\ref{fig:detection_score}. Based on the results, we can justify our design of $\Delta$: In the left column, we can see the confidence of correctly classified examples is very high, but the wrong ones show uniform distribution. Conversely, in the middle column, although MLLM poses slightly lower scores on correct ones, it significantly suppresses the confidence of wrong ones. As a result, we combine two scores to obtain $\Delta$, which can produce clearly separable distributions to benefit the diagnosing process. Unless specified, we set the threshold $\delta=0.6$ which works effectively in most scenarios.

\begin{table}[t]
\small
\centering
\vspace*{-4mm}
\caption{\small Comparison of classification accuracy (\%) on 5 OOD datasets with Otter~\cite{li2023otter} and MMICL~\cite{zhao2023mmicl}. We compare the performance on CLIP ViT-L~\cite{radford2021learning} backbone.}
\vspace{1mm}
\setlength{\tabcolsep}{1.5mm}
\begin{tabular}{l|l|ccccc}
\toprule
MLLM & Method & IN-A & IN-R & IN-SK & IN-V & iWildCam \\
\hline\hline
None & CLIP & 69.3 & 86.6 & 59.4 & 51.8 & 13.4 \\

% & EVA~\cite{EVA} & 71.9 & 91.4 & 67.7 & 64.9 & 21.9 \\
\hline
\multirow{2}{*}{\shortstack{Otter}}
     & MVT & 64.1 & 85.2 & 59.5 & 51.9 & \bf16.2 \\
     & +FT & \bf73.5 & \bf88.7 & \bf60.0 & \bf55.7 & -  \\
\hline
\multirow{2}{*}{\shortstack{MMICL}} 
     & MVT & 71.2 & 88.1 & 59.0 & 62.1 & \bf25.0 \\
     & +FT & \bf75.1 & \bf89.5 & \bf61.4 & \bf68.8 & - \\
\bottomrule
\end{tabular}
\label{tab:mllm_model}
\vspace{-6mm}
\end{table}

\vspace{-3mm}
\paragraph{Ablation Study on MLLM Backbone.}
To testify the effectiveness of MVT on different MLLM backbones, here we instantiate our method using Otter~\cite{li2023mimicit} and compare it to the previous realization on MIMIC~\cite{zhao2023mmicl}. The result is shown in Table~\ref{tab:mllm_model}. We can see that both the implementation on MMICL and Otter show superior performance to the employed vision encoder backbone. Although the performance slightly differs between Otter and MMICL, which could be due to the model capacity and their training strategy, we can generally conclude that our MVT method is applicable to different MLLMs backbones with ICL and could further benefit from more sophisticated MLLMs in the future.

\vspace{-2mm}
\subsection{Performance Analysis}
\vspace{-2mm}
\label{sec:performance_analysis}
Further, we conduct qualitative analysis to thoroughly validate the effectiveness of our MVT.

\begin{figure}[t]
\vspace{-2mm}
\begin{minipage}[h]{0.23\textwidth}
    \centering
    \includegraphics[width=\linewidth]{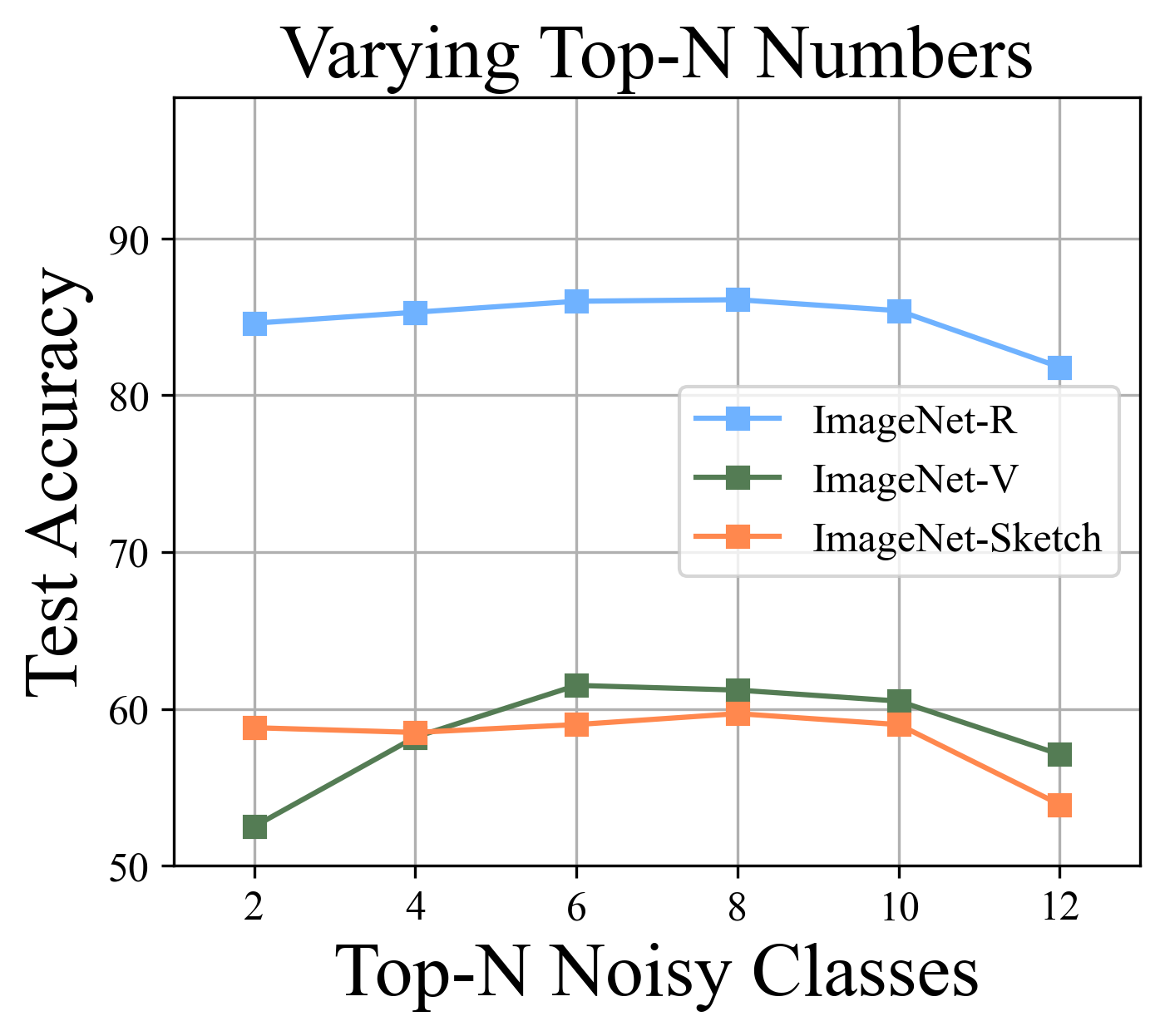}
\vspace{-9mm}
    \caption{\small Performance analysis by varying the number of top-$N$ chosen noisy classes.}
    \label{fig:vary_top_n}
\end{minipage}\ \ \ 
\begin{minipage}[h]{0.23\textwidth}
    \centering
    \includegraphics[width=\textwidth]{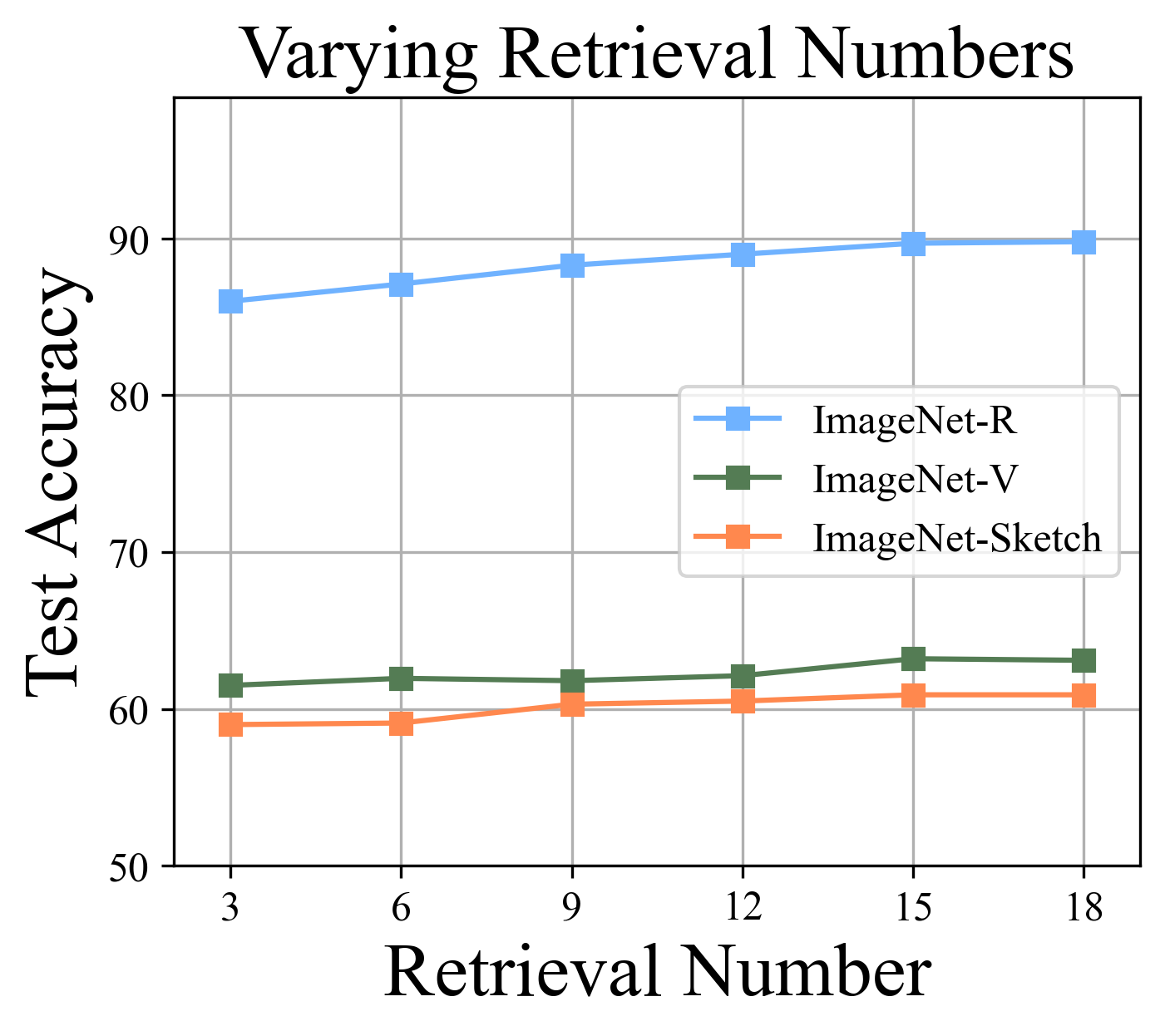}
\vspace{-9mm}
    \caption{\small Performance analysis by varying the number of retrieved exemplars.}
    \label{fig:vary_retrieve_number}
\end{minipage}
\vspace{-5mm}
\end{figure}

\vspace{-2mm}
\paragraph{Choice of Top-$N$ Noisy Classes.}
To study how a varied number of chosen noisy classes could affect the performance of our method, we change the top-$N$ number from $2$ to $12$, and show the result on ImageNet-R, ImageNet-V, and ImageNet-Sketch datasets in Figure~\ref{fig:vary_top_n}. We find a common phenomenon that either too small or too large a number of $N$ could hurt the performance. This could be because that small $N$ would ignore too many potential ground-truth classes. In contrast, large $N$ includes too many choices that could interfere with the final prediction. Setting $N$ to $6$ could be an ideal choice for ImageNet-based datasets.

\vspace{-2mm}
\paragraph{Effect of Retrieval Numbers.}
In our experiments, we retrieve exemplars for $3$ times and average the predictions. To further investigate the effect of varied retrieval numbers, we change the number of retrievals from $3$ to $18$ and conduct experiments on the same OOD datasets as above. Specifically, we consider one positive and negative pair for a single DICL round as one retrieval. We repeat this process for $R$ times and ensemble the MLLM predictions through $\frac{1}{R}\sum_r^R\left[z_c[\text{True}]^r, z_c[\text{False}]^r\right]$. In this way, it is possible that MLLM predictions would be more accurate. The result is shown in Figure~\ref{fig:vary_retrieve_number}. We observe that the performance steadily improves as the retrieval number increases, however, the performance gains vanish when the retrieval number becomes too large. Moreover, large retrieval numbers would multiply the computation cost. Therefore, it is suggested to set the number to a reasonably small value.

\vspace{-3mm}
\paragraph{Performance of Different Retrieval Strategy.}
\setlength{\intextsep}{3pt}
\setlength{\columnsep}{4pt}
\begin{wrapfigure}{r}{4cm}
\vspace{-1mm}
    \centering
    \includegraphics[width=\linewidth]{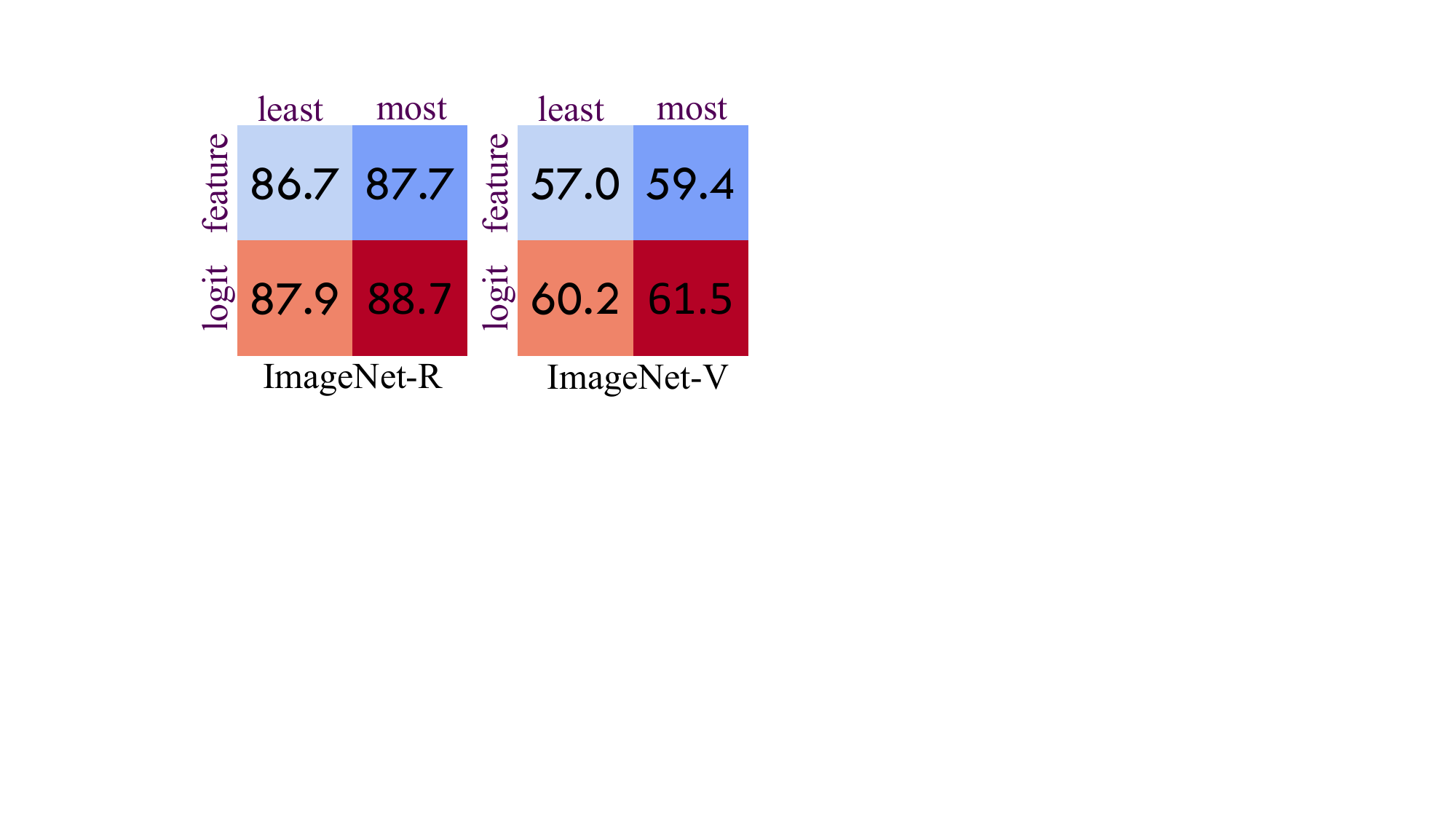}
    \vspace{-8mm}
    \caption{\small Performance analysis on different retrieval strategies.}
    \label{fig:retrieve_strategy}
\end{wrapfigure}
As shown by Alayrac et al.~\cite{alayrac2022flamingo}, Retrieval-based In-Context Example Selection (RICES) can significantly affect the ICL performance. Therefore, here we investigate its influence. Specifically, we propose two retrieval strategies, namely feature-based retrieval and logit-based retrieval. The former one is based on feature similarity and the latter one is based on the prediction logit. For each strategy, we conduct experiments on selecting the most similar examples and the least similar examples, which are denoted as ``most'' and ``least'', respectively. The results are shown in Figure~\ref{fig:retrieve_strategy}. Apart from the intuitive finding that least-similar retrieval is inferior to selecting the most-similar one, we also observe that logit-based retrieval is more effective than feature-based one. We assume this is due to the image classification task is more related to logit value rather than feature similarity.

\begin{figure}[t]
\vspace{-0mm}
\begin{minipage}[h]{0.23\textwidth}
    \centering
    \includegraphics[width=\linewidth]{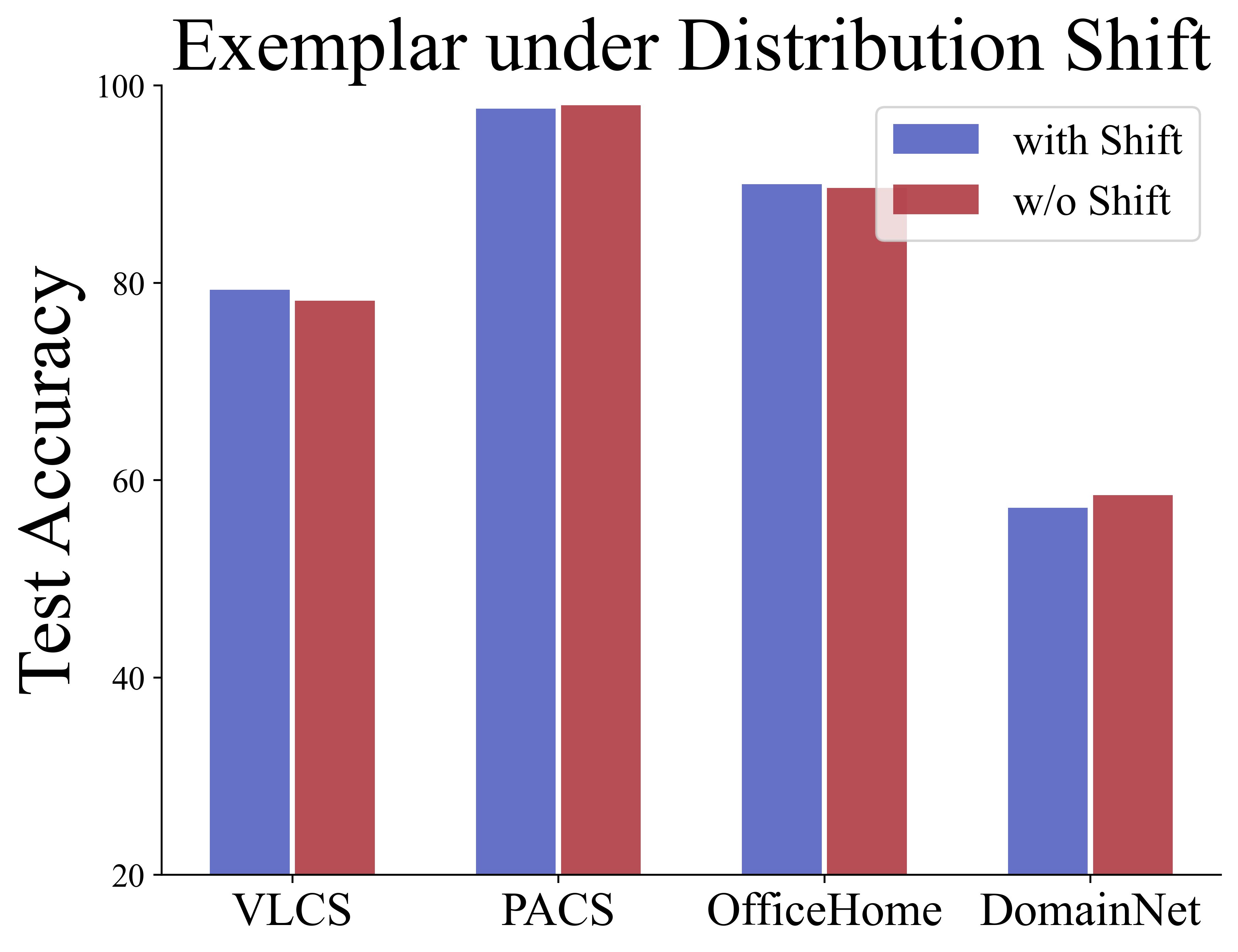}
    \vspace{-7mm}
    \caption{\small Performance analysis on in-context exemplars with distribution shift.}
    \label{fig:dicl_distshift}
\end{minipage}\ \ \ \begin{minipage}[h]{0.23\textwidth}
    \centering
    \includegraphics[width=\textwidth]{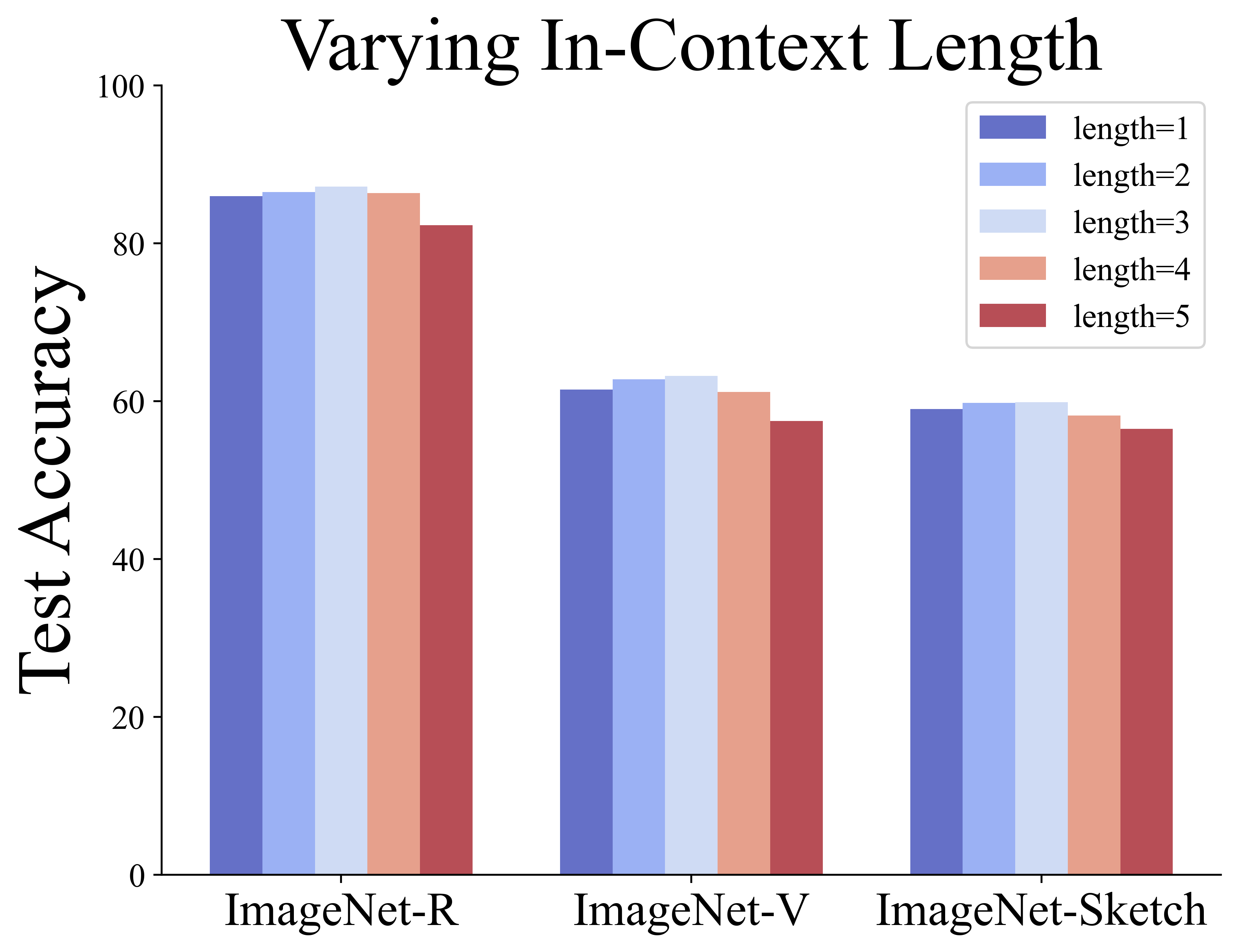}
    \vspace{-7mm}
    \caption{\small Performance analysis by varying the length of in-context exemplars.}
    \label{fig:dicl_length}
\end{minipage}
\vspace{-5mm}
\end{figure}

\vspace{-3mm}
\paragraph{Effect of In-Context Exemplars with Distribution Shift.}
When the support set suffers from a distribution shift from the target OOD dataset, whether DICL can still perform robustly remains to be validated. Hence, we leave one domain out as our support set and leverage the rest domains as our target OOD dataset. In comparison, we choose a small hold-out data split as the support set which shares the same distribution as the OOD dataset. The results are shown in Figure~\ref{fig:dicl_distshift}. Surprisingly, we find that the performance is not influenced by the distribution shift, which demonstrates that our MVT can still be effective when exemplars are retrieved from different distributions.

\vspace{-2mm}
\paragraph{Effect of Varying In-Context Length.}
Further, we analyze the effect of increasing exemplar length during inference. Particularly, we consider one positive and negative exemplar pair as length $1$. Here we vary the length from $1$ to $5$ and show the results in Figure~\ref{fig:dicl_length}. We observe slight improvement when the length gradually increases which is consistent with the theoretical findings~\cite{xie2021explanation}. However, when the length is longer than $4$ the performance drops and the predictions of MLLM become unstable which could be other than ``True'' or ``False''. This might be due to the limited capacity of MLLMs on handling a certain amount of information, which is worth conducting studies on sophisticated MLLMs in the future.

\begin{figure}[t]
\vspace{0mm}
\centering
\includegraphics[width=\linewidth]{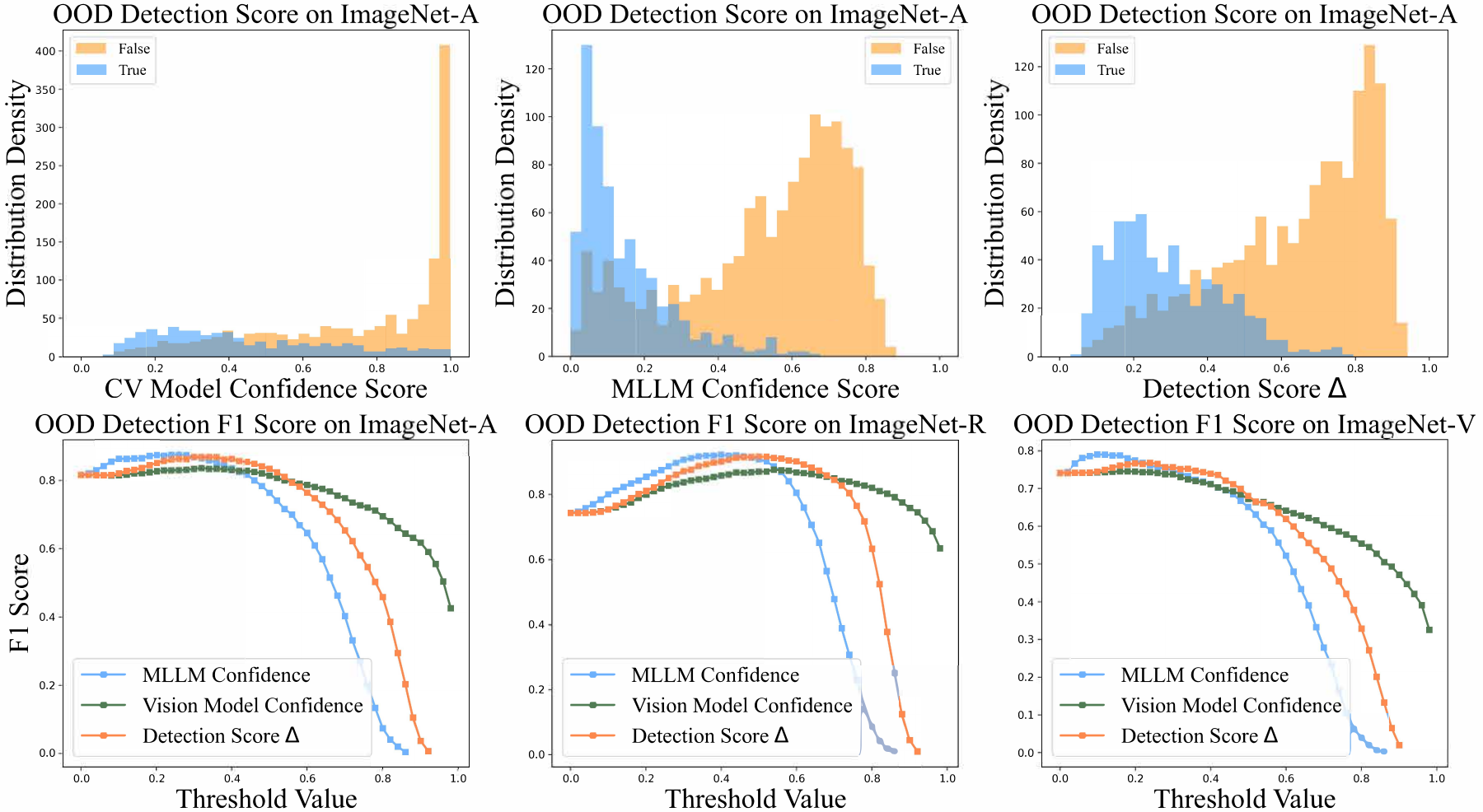}
\vspace*{-3mm}  
\caption{\small OOD detection analysis. Upper: Detection score distribution on ImageNet-A; Lower: F1 scores of vision model confidence, MLLM diagnosing confidence, and our $\Delta$ score in ImageNet-A, ImageNet-R, and ImageNet-V.}
\label{fig:ood_detection}
\vspace*{-2mm}
\end{figure}

\vspace{-2mm}
\paragraph{Performance of OOD Detection.}
At last, we consider a more challenging scenario where data from open classes could exist in the target dataset. Here we simulate this situation by choosing $60\%$ of the classes as closed classes and the rest are open classes. To detect such open-class data, \textit{i.e.}, OOD detection~\cite{hendrycks2016baseline, wang2024learning}\footnote{Note that OOD detection here is different from the previous setting: here we focus on detecting open-class data, and previous one focuses on detecting prediction errors.}, we use the vision model prediction confidence as a baseline and compare it with the MLLM diagnosing confidence as well as the detection score $\Delta$ in Eq.~\ref{eq:detection_score}. The result is shown in Figure~\ref{fig:ood_detection}. In the upper row, we observe the similar clearly distinguishable distributions using our score $\Delta$ as in Figure~\ref{fig:detection_score}. In the lower row, we show the F1 score of each detection criterion under a threshold varied from $0$ to $1$ on three datasets. When a criterion produces confidence larger than the threshold, it would predict as close-class data, other as open-class ones. Based on the result, we find that MLLM achieves better detection performance when the threshold is small, but vision model confidence is relatively better when the threshold is large, \textit{i.e.}, MLLM can effectively detect open classes while vision models are better at recognizing close classes. However, an effective detection should have a reasonable threshold value that is neither too large nor too small and meanwhile has a high F1 score. Hence, by combining them together, our detection score $\Delta$ can achieve the best F1 score when the threshold is around the middle range.

\setlength{\intextsep}{1.4pt}
\setlength{\columnsep}{2pt}

%% file: sec/5_conclusion.tex
\vspace{-2mm}
\section{Conclusion}
\vspace{-2mm}
In this paper, we propose a novel paradigm of fine-tuning vision models via leveraging MLLMs to improve visual robustness on downstream OOD tasks. Specifically, we effectively estimate a transition matrix to help find the most probable noisy classes. By using a positive exemplar and a negative exemplar retrieved based on the noisy classes, we can conduct DICL to rectify incorrect vision model predictions through two stages dubbed diagnosing and therapy. Thanks to the rectified predictions, the robustness of vision models can be further improved through fine-tuning. We conduct detailed theoretical analysis and extensive quantitative and qualitative experiments to justify the proposed method. Our framework can significantly reduce the cost of training vision models and provide insights into many visual recognition problems such as OOD detection, OOD generalization, weakly-supervised learning, \textit{etc}.

\clearpage

\section*{Acknowledgements}
Yinpeng Dong is supported by NSFC project (No.~62276149). Hang Su is partially supported by NSFC projects (Nos.~92248303, 92370124, 62350080). Shibao Zheng is supported by NSFC projects (Nos.~62071292, U21B2013) and STCSM project (No. 22DZ2229005). Tongliang Liu is partially supported by the following Australian Research Council projects: FT220100318, DP220102121, LP220100527, LP220200949, and IC190100031.

\section*{Impact Statement}
As the rapid development of MLLMs and LLMs has started influencing our lives, the proposed work could also show some potential broader impacts. For example, when highly powerful and accurate MLLMs occur in the future, human supervision for learning models will be no longer necessary. The learning paradigm could be large models guiding small models. However, it is vital that researchers make sure the knowledge interaction between large models and small models is ethically responsible, trustworthy, secure, robust, and explainable. Harmful knowledge or uncertain knowledge should not be freely transferred from one model to another. Hence, a rigorous code of ethics should be formulated to effectively regulate the process.

%% file: sec/X_suppl.tex
\onecolumn 
\renewcommand*\contentsname{\centering{Index}}

\vspace{-3mm}
\begin{center}
	\rule{6.875in}{1.2pt}\\ % 4.0
	{\Large\bf Supplementary Material for\\ ``Machine Vision Therapy: Multimodal Large Language Models\\Can Enhance Visual Robustness via Denoising In-Context Learning''}
	\rule{6.875in}{1.2pt}
\end{center}
% \begingroup
% \color{black}
% \tableofcontents
% \endgroup
% \newpage

In this supplementary material, we provide extensive quantitative and qualitative studies on a wide range of datasets to thoroughly understand the essence of the proposed framework. First, we discuss some related works on OOD generalization and Multimodal Large Language Models in Section~\ref{sec:app_related_work}. Second, we theoretically analyze and justify the proposed DICL strategy in Section~\ref{sec:app_theoretical_framework}. Then, we elucidate the additional details of our experimental setting and implementation in Section~\ref{sec:app_details}. Further, we provide additional quantitative comparisons on different MLLM and CV backbone models, and different OOD types in Section~\ref{sec:app_comparisons}. Finally, we perform additional analysis to further explore the effectiveness of our method in Section~\ref{sec:app_analysis}. Finally, we discuss the limitation of this work and the broader impact in Section~\ref{sec:app_limitation}

\section{Related Work}
\label{sec:app_related_work}

In this section, we provide a brief discussion of the OOD generalization problem and multimodal large-language models.
\subsection{OOD Generalization}
OOD data refers to those with different distributions from training data. OOD generalization aims at improving the performance of deep models to unseen test environments. Researchers attempted to tackle the problem from different perspectives, such as data augmentation, OOD detection, invariant causal mechanisms~\cite{huang2023harnessing, huang2023winning, zhou2021towards, zhou2022improving}, and so on. Data augmentation is effective in improving model generalization. Typical methods involve Cutout~\cite{devries2017improved}, which randomly occludes parts of an input image; CutMix~\cite{yun2019cutmix}, which replaces a part of the target image with a different image; Mixup~\cite{zhang2017mixup}, which produces a convex combination of two images; DeepAugment~\cite{hendrycks2021many}, which passes a clean image through an image-to-image network and introduces several perturbations during the forward pass. Some methods conduct OOD detection to separate OOD data. Typical methods include softmax confidence score~\cite{hendrycks2016baseline, huang2023robust}, which is a baseline for OOD detection; Outlier Exposure (OE)~\cite{hendrycks2018deep}, which uses unlabeled data as auxiliary OOD training data. Energy scores are shown to be better for distinguishing OOD samples from IID ones~\cite{liu2020energy}. Some work resort to causality to study the OOD generalization problem. Typical methods include MatchDG~\cite{mahajan2021domain}, which proposes matching-based algorithms when base objects are observed and approximate the objective when objects are not observed; INVRAT~\cite{chang2020invariant}, which leveraged some conditional independence relationships induced by the common causal mechanism assumption.

\subsection{Multimodal Large Language Models}
The field of vision-language models has witnessed significant advancements in recent years, driven by the growing synergy between computer vision and natural language processing. Notably, this synergy has led to the exceptional zero-shot performance~\cite{hou2024visual} of CLIP~\cite{radford2021learning}, a model that employs a two-tower contrastive pretraining approach to align image and text information. In the rapidly evolving landscape of LLMs, exemplified by GPTs~\cite{brown2020language}, LLaMA~\cite{touvron2023llama}, and Vicuna~\cite{vicuna2023}, it has become evident that LLMs possess the capacity to process information from diverse domains. BLIP-2~\cite{li2023blip}, for instance, serves as a foundational model, aligning visual features and text features using a Querying Transformer (Q-former) and utilizing OPT~\cite{zhang2022opt} and FLAN~\cite{chung2022scaling} as language models. Building upon BLIP-2, Instruct-BLIP~\cite{dai2023instructblip} has enhanced instruction-following capabilities. To further bolster the instruction-following proficiency of multi-modal models, LLaVA~\cite{liu2023llava} and Mini-GPT4~\cite{zhu2023minigpt} have introduced meticulously constructed instruction sets, which have found widespread application in various multi-modal models. mPLUG-Owl~\cite{ye2023mplug} introduces a two-stage learning paradigm, first fine-tuning the visual encoder and then refining the language model with LoRA~\cite{hu2021lora}. This approach effectively fuses image and text features. Some models consider additional modalities, such as ImageBind~\cite{girdhar2023imagebind}, which simultaneously incorporates data from six modalities without the need for explicit supervision, and PandaGPT~\cite{su2023pandagpt} which enhances its instruction-following capabilities. Several multi-modal models prioritize the in-context learning abilities of LLMs. Flamingo~\cite{alayrac2022flamingo}, in one of the pioneering efforts, integrates a gated cross-attention module to align with the spaces of images and text. Otter~\cite{li2023otter} refines OpenFlamingo~\cite{awadalla2023openflamingo}, an open-source version of Flamingo, by introducing instruction-tuning datasets to improve instruction-following abilities. Multi-Modal In-Context Learning (MMICL)~\cite{zhao2023mmicl} is a comprehensive vision-language model that incorporates Instruct-BLIP, enabling the analysis and comprehension of multiple images, as well as the execution of instructions. MLLMs possess the remarkable capacity to capture intricate details and engage in reasoning when presented with an image. Nevertheless, it remains uncertain about how to enhance visual perception by harnessing the knowledge embedded within LLMs.

\section{Theoretical Framework}
\label{sec:app_theoretical_framework}
\paragraph{Pretraining distribution formulation.} We based on the in-context learning framework proposed by~\citet{xie2021explanation}. In this framework, a latent concept $\phi$ from a concept space $\Phi$ defines a pretraining distribution $p$ over prompt tokens $o$ observed from a vocabulary space $O$. To generate the desired content, we first sample a concept from a prior $p(\phi)$ and then sample the tokens conditioned on the concept. We denote the total length of the pretraining examples is $T$:
\begin{equation}
    p(o_1, \ldots, o_T)=\int_{\phi\in\Phi} p(o_1, \ldots, o_T|\phi)p(\phi)d\phi.
\end{equation}
The conditional probability $p(o_1, \ldots, o_T|\phi)$ is defined by a Hidden Markov Model (HMM). Based on the concept $\phi$, a transition matrix of the HMM hidden states $h_1, \ldots, h_T$ from a hidden state space $\mathcal{H}$ can be found.

\paragraph{Prompt distribution formulation.} During the in-context learning process, we sample a prompt from a new distribution $p_{prompt}$, which contains $n$ independent exemplars and 1 query example, which are all conditioned on a shared prompt concept $\phi^*$. The goal is to predict the next token based on the exemplars and the query example. Specifically, the $i$-th exemplar $O_i$ consists of a tokenized image feature $x_i=O_i[1:k_x]$, a text description to claim the class of the image $y_i=O_i[k_x:k_x+k_y]$, and a binary prediction to judge whether the claim of the image is ``True'' or ``False'', which is denoted by $z_i=O_i[k_x+k_y:k_x+k_y+1]$ at the end of each exemplar. The generating process of the $i$-th exemplar is as follows: 
\begin{enumerate}
    \item First generate a start hidden state $h_i^{start}$ from prompt start distribution $p_{prompt}$;
    \item Given $h_i^{start}$, generate the exemplar sequence $O_i=[x_i, y_i, z_i]$ from $p(O_i|h_i^{start}, \phi^*)$, the generate exemplars are conditioned on a given prompt concept $\phi^*$.
\end{enumerate}

The query example $Q$ is sampled similarly without the binary prediction $z_q$, \textit{i.e.} $Q=[x_q, y_q]$. Between each exemplar and the query example, there is a special delimiter token $o^{delim}$ denoting the end of each exemplar sequence. The prompt can be formulated as follows:
\begin{equation}
    [S_n, Q] = [x_1, y_1, z_1, o^{delim}, x_2, y_2, z_2, o^{delim}, \ldots, x_n, y_n, z_n, o^{delim}, x_q, y_q]\sim p_{prompt},
\end{equation}
where $S_n$ denotes the $n$ independent exemplars for in-context demonstration.

\paragraph{Denoising In-context learning task.} In our denoising in-context learning, the output prediction $z$ for each image and text pair $[x, y]$ is sampled based on the prompt distribution $p_{prompt}(z|x, y)$:
\begin{equation}
    z_q\sim p_{prompt}(z|x, y)=\mathbb{E}_{h_q^{start}\sim p_{prompt}(h_q^{start}|Q)}[p(z|Q, h_q^{start}, \phi^*)],
\end{equation}
where $h_q^{start}$ denotes the hidden state corresponding to the first token of $Q$, \textit{i.e.,} the first token of $x_q$.

Our goal is to analyze the in-context predictor $f_n(x_q, y_q) = \arg\max_{z}p(z|S_n, x_q, y_q)$ which outputs the most likely prediction over the pretraining distribution $p$ conditioned on the exemplars $S_n$ sampled from the prompt distribution $p_{prompt}$. We assume the in-context predictor is trained by $0-1$ error with $n$ training examples $\mathcal{L}_{0-1}(f_n)=\mathbb{E}_{x_q, y_q}\sim p_{prompt}[\mathbf{1}[f_n(x_q)\neq y_q]]$ and $\mathcal{L}_{0-1}(f_n)=\mathbb{E}_{x_q, y_q, z_q}\sim p_{prompt}[\mathbf{1}[f_n(x_q, y_q)\neq z_q]]$..

One major difference of our denoising in-context learning strategy is that we not only use positive exemplars that show exact image-text match, \textit{i.e.,} $(x, y)\sim p(x, y)=p_{prompt}(x, y)$, we also have negative exemplars where image and text are not corresponding to each other. To construct such a prompt, we have to first select the ideal $y$, based on the matching result of $x$ and $y$, we can further obtain the prediction $z$. Therefore, in the following theoretical proof, we propose to conduct two-step analyses on $y$ and $z$, respectively. 

\subsection{Assumptions}
Our theoretical framework is built upon~\citet{xie2021explanation}, whose assumptions are also applied to our analysis.

\begin{assumption}[Delimiter hidden states]
\label{ass:delimiterstates}
Let the delimiter hidden states $\mathcal{D}$ be a subset of $\mathcal{H}$. For any $h^{delim}\in \mathcal{D}$ and $\phi \in \Phi$, $p(o^{delim} \vert h^{delim}, \phi)=1$ and for any $h\notin \mathcal{D}$, $p(o^{delim} \vert h, \phi)=0$.
\end{assumption}

\begin{assumption}[Bound on delimiter transitions]
\label{ass:delimiterbound}
For any delimiter state $h^{delim} \in \mathcal{D}$ and any hidden state $h \in \mathcal{H}$, the probability of transitioning to a delimiter hidden state under $\phi$ is upper bounded $p(h^{delim} \vert h, \phi) < c_2$ for any $\phi \in \Phi \setminus \{\phi^*\}$, and is lower bounded $p(h^{delim} \vert h, \phi^*) > c_1 > 0$ for $\phi^*$.
Additionally, the start hidden state distribution for delimiter hidden states is bounded as $p(h^{delim} \vert \phi) \in [c_3, c_4]$.
\end{assumption}

\begin{assumption}[Distribution shift from prompt start distribution]
\label{ass:promptstartshift}
We assume that the prompt start distribution $p_{prompt}$ is close in TV distance to all hidden transition distributions (under $\phi^*$) starting from a delimiter hidden state: $\max_{h^{delim} \in \mathcal{D}} TV(p_{prompt}(h) \| p(h \vert h^{delim}, \phi^*)) < \tau / 4$.
Here, $\tau = p_{prompt}(y_{max} \vert Q) - \max_{y \neq y_{max}}p_{prompt}(y \vert Q)$ is the margin between the most likely label $y_{max} = \arg\max_{y} p_{prompt}(y \vert Q)$ and the second most likely label.
\end{assumption}

\begin{assumption}[Well-specification]
\label{ass:wellspecified}
The prompt concept $\phi^*$ is in $\Phi$.
\end{assumption}

\begin{assumption}[Regularity]
\label{ass:regularity}
The pretraining distribution $p$ satisfies: 1) Lower bound on transition probability for the prompt concept $\phi^*$: for any pair of hidden states $h, h' \in \mathcal{H}$, $p(h \vert h', \phi^*) > c_5 > 0$. 
2) Start hidden state is lower bounded: for any $h \in \mathcal{H}$, $p(h \vert \phi^*)\geq c_8 > 0$.
3) All tokens can be emitted: for every symbol $o$, there is some hidden state $h\in \mathcal{H}$ such that $p(o \vert h, \phi^*) > c_6 > 0$,
4) The prior $p(\phi)$ has support over the entire concept family $\Phi$ and is bounded above everywhere.
\end{assumption}

Except from the above five adapted assumptions from~\citet{xie2021explanation}, we have an another mild assumption:

\begin{assumption}[Distribution consistency]
\label{ass:consistency}
The pretraining distribution $p$ and prompt distribution $p_{prompt}$ satisfy $\forall (x_q, y_q)\sim p_{prompt}, p(x_q, y_q)=p_{prompt}(x_q, y_q)$.
\end{assumption}
This assumption indicates that the chosen prompt distribution is a sub-distribution of the pretraining distribution and the joint distribution of $x_q$ and $y_q$ is consistent across $p$ and $p_{prompt}$. This assumption avoids the situations where there are concept shifts between $p$ and $p_{prompt}$, \textit{i.e.}, all $y\sim p_{prompt}$ are known in $p$ and can find an exact match for each $x_q$ in $p$.

\subsection{Theoretical Proof}
We first show that given a query image $x_q$, when conditioned on a concept $\phi^*$ and prompt $S_n$, the most probable text output token for $y_q$ is the same as the class in the prompt distribution $p_{prompt}$ with maximum probability. Then, we show that: in our denoising in-context learning, when achieving the most likely prediction $z$ output by the MLLM predictor, the class text $y_q$ in the pretraining distribution $p$ is the same as the one found by the prompt distribution $p_{prompt}$, which is the exact match for the give image $x_q$.

Before we start analyzing the binary prediction $z$, we first investigate the most probable class $y$ given prompt and query image $\arg\max_y p(y\vert S_n, x_q)$. 
\begin{theorem}
    Assume that the above assumptions hold, if for all $\phi\in\Phi$, $\phi\neq\phi^*$, the concept $\phi^*$ satisfies the distinguishability condition: $\sum_{j=1}^k KL_j(\phi^*\|\phi) > \epsilon_{start}^{\phi} + \epsilon_{delim}^{\phi}$, then as $n\rightarrow\infty$, the prediction $y$ according to the pretraining distribution is
    \begin{equation}
        \arg\max_y p(y\vert S_n, x_q, \phi^*) \rightarrow \arg\max_y p_{prompt}(y\vert x_q).
    \end{equation}
    Thus, the in-context predictor $f_n$ achieves the optimal $0-1$ risk: $\lim_{n\rightarrow\infty}\mathcal{L}_{0-1}(f_n)=\inf_f\mathcal{L}_{0-1}(f)$.
    \label{theo:optimal_y_supp}
\end{theorem}
The detailed proof of this theorem is similar to~\citet{xie2021explanation}, please refer to the Section D of the original paper.

Under this assumption, the in-context predictor is guaranteed to have the highest probability of generating the class description $y$ that exactly matches the query image $x_q$. In another way, when $x_q$ does not belong to $y$, the probability $p(y\vert S_n, x_q)$ is less than the optimal value.

\begin{lemma}
    Under the same condiction of Theorem~\ref{theo:optimal_y_supp}, the prediction $z$ according to the pretraining distribution is
    \begin{equation}
        \arg\max_z p(z\vert S_n, x_q, y_q, \phi^*) \rightarrow \arg\max_z p_{prompt}(z\vert x_q, y_q).
    \end{equation}
    \label{lemm:optimal_z_supp}
\end{lemma}

Lemma~\ref{lemm:optimal_z_supp} can be easily derived based on Theorem~\ref{theo:optimal_y_supp} by considering $y$ as a fixed prompt in $Q$.

\begin{theorem}
    Assume that the above assumptions hold, as $n\rightarrow\infty$, when achieving the largest prediction probability of $z$ given prompt under concept $\phi^*$, the corresponding class description $y$ follows the same $y$ obtained from the prompt distribution:
    \begin{equation}
        \arg\max_y p(z\vert S_n, x_q, y, \phi^*) \rightarrow \arg\max_y p_{prompt}(z\vert x_q, y)
    \end{equation}
    \label{theo:dicl_supp}
\end{theorem}
\begin{proof}
    Since we already have Theorem~\ref{theo:optimal_y_supp}, if we can prove that $\arg\max_y p(y\vert S_n, x_q, \phi^*)=\arg\max_y p(z\vert S_n, x_q, y, \phi^*)$ and $\arg\max_y p_{prompt}(y\vert x_q)=\arg\max_y p_{prompt}(z\vert x_q, y)$, then we can complete the justification.
    \begin{align}
        p(z\vert S_n, x_q, y, \phi^*)=\sum_{h^{start}_q\in \mathcal{H}}p(z\vert h^{start}_q)p(h^{start}_q\vert S_n, x_q, y, \phi^*).
    \end{align}
    By expanding the last term, we have:
    \begin{align}
        p(h^{start}_q\vert S_n, x_q, y, \phi^*)=&\frac{p(x_q, y\vert h^{start}_q, S_n, \phi^*)p(h^{start}_q)}{p(x_q, y)}\\
        \propto& \frac{p(x_q, y\vert h^{start}_q, S_n, \phi^*)}{p(x_q, y)}
    \end{align}
    where $p(h^{start}_q)$ is considered as a constant. Moreover, based on Assumption~\ref{ass:consistency}, the joint distribution $p(x_q, y)$ is predefined by the pretraining distribution, which does not affect the marginal distribution of $z$, thus we can have
    \begin{align}
        \frac{p(x_q, y\vert h^{start}_q, S_n, \phi^*)}{p(x_q, y)}&=\frac{p(y\vert S_n, x_q, h^{start}_q, \phi^*)p(x_q\vert h^{start}_q)}{p(x_q, y)}\\
        &\propto p(y\vert S_n, x_q, h^{start}_q, \phi^*)p(x_q\vert h^{start}_q).
    \end{align}
    Since the change of $y$ does not affect the quantity of $p(z\vert h^{start}_q)$, therefore, applying argmax on both sides of the equation holds for finding the optimal $y$:
    \begin{align}
        \arg\max_y p(z\vert S_n, x_q, y, \phi^*)&=\arg\max_y \sum_{h^{start}_q\in \mathcal{H}}p(z\vert h^{start}_q)p(y\vert S_n, x_q, h^{start}_q, \phi^*)\\
        &=\arg\max_y p(y\vert S_n, x_q, h^{start}_q, \phi^*).
    \end{align}
    Similarly,
    \begin{align}
        p_{prompt}(z\vert x_q, y)&= \sum_{h^{start}_q\in \mathcal{H}}p(z\vert h^{start}_q, \phi^*)p_{prompt}(h^{start}_q\vert x_q, y),\\
        p_{prompt}(h^{start}_q\vert x_q, y)&= \frac{p_{prompt}(x_q, y\vert h^{start}_q)p_{prompt}(h^{start}_q)}{p_{prompt}(x_q, y)}\\
        &\propto p_{prompt}(x_q, y\vert h^{start}_q)\\
        &\propto p_{prompt}(y\vert x_q, h^{start}_q)p_{prompt}(x_q\vert h^{start}_q),\\
        \arg\max_y p_{prompt}(z\vert x_q, y) &= \arg\max_y \sum_{h^{start}_q\in \mathcal{H}}p_{prompt}(z\vert h^{start}_q, \phi^*)p_{prompt}(y\vert x_q, h^{start}_q)\\
        &=\arg\max_y p_{prompt}(y\vert x_q, h^{start}_q),
    \end{align}
    where the change of $y$ still does not affect the quantity of $p_{prompt}(z\vert h^{start}_q, \phi^*)$. Since
    \begin{align}
        p(y\vert S_n, x_q, \phi^*)&=\sum_{h^{start}_q\in \mathcal{H}}p(y\vert h^{start}_q, S_n, x_q, \phi^*)p(h^{start}_q\vert S_n, x_q, \phi^*),\\
        p_{prompt}(y\vert x_q)&=\sum_{h^{start}_q\in \mathcal{H}}p_{prompt}(y\vert h^{start}_q, x_q)p_{prompt}(h^{start}_q\vert x_q),
    \end{align}
    it is easy to find that
    \begin{align}
        \arg\max_y p_{prompt}(y\vert x_q, h^{start}_q) = \arg\max_y p_{prompt}(y\vert x_q),\\
        \arg\max_y p(y\vert S_n, x_q, h^{start}_q, \phi^*) = \arg\max_y p(y\vert S_n, x_q, \phi^*).
    \end{align}
    Thus, we have that as $n\rightarrow\infty$,
    \begin{equation}
        \arg\max_y p(z\vert S_n, x_q, y, \phi^*) \rightarrow \arg\max_y p_{prompt}(z\vert x_q, y).
    \end{equation}
\end{proof}

Lemma~\ref{lemm:optimal_z_supp} and Theorem~\ref{theo:dicl_supp} together show that when given a query image $x_q$, if the chosen query class description $y_q$ is the true class of $x_q$, then under the given assumptions, the binary prediction $z$ for judging the correctness of the image-text pair would be the maximum value compared to all other class descriptions $y\neq y_q, y\in \mathcal{Y}$. Therefore, we can justify that using an in-context predictor can help identify the true class label of a given image.

\section{Additional Details}
\label{sec:app_details}
We run all our experiments on 8 NVIDIA 3090 GPUs using the Pytorch framework. During MVT, we freeze the MLLM backbone model to stop generating gradients that might cause additional computational costs. Then, for the vision encoder, we use \code{model.eval()} to produce vision predictions. Additionally, the predictions are evaluated and corrected. Based on the rectified predictions, we use \code{torch.optim.Adam()} or \code{torch.optim.SGD()} optimizer to fine-tune the vision model for $3$ epochs. Note that we conduct fair comparisons in each experiment by using the same optimizer. Due to the memory of ViT-g is larger, thus we use \code{torch.optim.SGD()} to optimize ViT-g model and \code{torch.optim.Adam()} for other vision models. The vision encoder is trained with \code{torch.float32} precision to prevent overflow. The batch size for ViT-L vision encoder is 16 and the batch size for ViT-g is $8$ with $2$ accumulation steps. The learning rate of the training process is $5e-7$ and the cosine scheduler for ViT-L with \code{torch.optim.Adam()}. Due to limited GPU memory, we fine-tune ViT-g with \code{torch.optim.SGD()} of learning rate $1e-4$ and $0$ momentum. Besides, experiments in Sec.~\ref{rn_vitb} in the Appendix do not follow the settings mentioned above. Because the vision encoders to be fine-tuned have a large capacity gap with the vision encoders in MLLMs. We need to fine-tune the vision encoder to match the performance of MLLMs. Therefore, the learning rate is adjusted to $1e-4$ and the training epoch is adjusted to $20$. Note that all the fine-tuned data from the evaluation dataset are the chosen ones for therapy. Then we test the performance of all baseline methods on a split-out test set.

In DICL, our prompt for multi-class classification tasks is as follows: 
\begin{tcolorbox}
	\code{This image \{replace\_token\} shows a photo of <\#text>, True or False; Answer:}
\end{tcolorbox}
where the \code{\{replace\_token}\} is further replaced by the image feature, and \code{<\#text>} is further replaced by the class name. The MMICL and Otter model we use can be found at \hyperlink{https://github.com/haozhezhao/mic}{https://github.com/haozhezhao/mic} and \hyperlink{https://github.com/Luodian/Otter}{https://github.com/Luodian/Otter}, respectively. All our fine-tuned vision models can be directly found in Openai CLIP models: \hyperlink{https://huggingface.co/openai}{https://huggingface.co/openai}.

\clearpage

\begin{table*}[h]
	\centering
	\caption{Classification accuracy (\%) of baseline CLIP models and our method with MMICL~\cite{zhao2023mmicl} and Otter~\cite{li2023otter} as the VLMs on 5 ID datasets and 5 OOD datasets. We compare the performance of our method, and the fine-tuned models supervised by our method with the baseline models, i.e., ViT-L from CLIP~\cite{radford2021learning}. Fine-tuning with both MMICL and Otter improves the classification accuracy.}
	\setlength{\tabcolsep}{1.85mm}
	\begin{tabular}{l|l|ccccc|ccccc}
		\toprule
		\multirow{2}{*}{MLLM}&\multirow{2}{*}{Method}&\multicolumn{5}{c|}{ID}&\multicolumn{5}{c}{OOD}\\
		\cline{3-12}
		&& IN-Val & IN-V2 & CIFAR10 & CIFAR100 & MNIST & IN-A & IN-R & IN-SK & IN-V & iWildCam \\
		\hline\hline
		% RN50 & \multirow{3}{*}{CLIP~\cite{radford2021learning}} & 59.7 & 52.6 & 71.5 & 41.9 & 58.5 & 23.9 & 60.7 & 35.4 & 31.1 & 8.2 \\
		% RN101 & & 61.7 & 56.2 & 80.8 & 48.8 & 51.6 & 30.2 & 66.7 & 40.9 & 35.4 & 12.3 \\
		% ViT-B & & 62.9 & 56.1 & 89.9 & 65.0 & 47.9 & 32.2 & 67.9 & 41.9 & 30.5 & 10.9 \\
		None & CLIP & 75.8 & 70.2& 95.6 & 78.2 & 76.4 & 69.3 & 86.6 & 59.4 & 51.8 & 13.4 \\
		\hline
		\multirow{2}{*}{\shortstack{MMICL}} & MVT & 75.2 & \bf70.8 & \bf97.9 & 78.9 & 53.0 & 71.2 & 88.1 & 59.0 & \underline{62.1} & \bf25.0 \\
		& +FT & \bf76.9 & \underline{70.5} & \underline{96.7} & \bf82.0 & \underline{79.2} & \bf75.1 & \bf89.5 & \bf61.4 & \bf68.8 & - \\
		\hline
		\multirow{2}{*}{\shortstack{Otter}} & MVT & 74.2 & 67.4 & 94.7 & 70.1 & 52.0 & 64.1 & 85.2 & 59.5 & 51.9 & \underline{16.2} \\
		& +FT & \underline{76.3} & 70.1 & 96.6 & \underline{81.8} & \bf81.3 & \underline{73.5} & \underline{88.7} & \underline{60.0} & 55.7 & - \\
		\bottomrule
	\end{tabular}
	\label{tab:app_otter_imagenet}
\end{table*}

\begin{table*}[h]
	\small
	\caption{Classification accuracy (\%) of baseline CLIP models and our method with MMICL~\cite{zhao2023mmicl} and Otter~\cite{li2023otter} as the VLMs on 4 subsets of DomainBed datasets, including VLCS, PACS, OfficeHome, and DomainNet. We compare the performance of our method and the fine-tuned models supervised by our method with the baseline models, i.e., ViT-L from CLIP~\cite{radford2021learning}. Fine-tuning with both MMICL and Otter improves the classification accuracy.}
	\vspace{-0mm}
	\centering
	\setlength{\tabcolsep}{1.28mm}
	\begin{tabular}{l|l|cccc|cccc|cccc|ccccc|c}
		\toprule
		\multirow{2}{*}{MLLM} & Datasets & \multicolumn{4}{c|}{VLCS} & \multicolumn{4}{c|}{PACS} & \multicolumn{4}{c|}{OfficeHome} & \multicolumn{5}{c|}{DomainNet} & \multirow{2}{*}{Avg} \\
		\cline{3-19}
		& method & 0 & 1 & 2 & 3 & 0 & 1 & 2 & 3 & 0 & 1 & 2 & 3 & 0 & 1 & 2 & 3 & 4 \\
		\hline
		\hline
		None & CLIP & 74.9 & 83.5 & 80.3 & 74.5 & \underline{97.8} & 97.4 & \underline{97.5} & \underline{99.4} & 87.7 & 92.7 & 85.7 & 85.6 & 61.1 & 62.1 & 60.2 & 78.4 & 51.1 & 80.6 \\
		\hline
		\multirow{2}{*}{\shortstack{MMICL}} & MVT & \underline{83.8} & \underline{89.0} & \underline{87.2} & \underline{80.3} & 97.6 & 97.5 & \bf98.0 & \underline{99.4} & 87.7 & \underline{93.4} & \underline{89.0} & \underline{88.5} & 61.3 & 62.1 & 60.4 & 78.7 & \underline{53.4} & \underline{82.8} \\
		
		& +FT & \bf84.2 & \bf89.8 & \bf87.9 & \bf82.5 & \bf98.0 & \bf98.2 & \bf98.0 & \bf99.8 & \bf90.9 & \bf95.0 & \bf90.9 & \bf90.8 & \bf62.5 & \bf63.8 & \bf62.4 & \bf80.1 & \bf54.0 & \bf84.0 \\
		\hline
		\multirow{2}{*}{\shortstack{Otter}} & MVT & 67.5 & 77.4 & 73.7 & 66.6 & 97.0 & 96.3 & 96.5 & 99.0 & 85.6 & 89.9 & 83.6 & 83.3 & 56.5 & 58.6 & 56.3 & 74.1 & 46.5 & 77.0 \\
		
		& +FT & 76.8 & 87.7 & 82.3 & 77.4 & \bf98.0 & \underline{97.7} & \bf98.0 & \bf99.8 & \underline{88.7} & \underline{93.4} & 87.7 & 87.1 & \underline{62.0} & \underline{63.0} & \underline{61.3} & \underline{79.7} & \underline{53.4} & 82.0 \\
		\bottomrule
	\end{tabular}
	\label{tab:app_otter_domainbed}
\end{table*}

\section{Additional Quantitative Comparisons}
\label{sec:app_comparisons}
Here we provide extensive quantitative comparisons of different MLLM and vision models, in various robustness settings.

\subsection{Quantitative Comparison using Otter}
First, similar to Table 1 in the main paper, here we conduct additional experiments on various ImageNet-based datasets and DomainBed datasets using ViT-L but a different MLLM backbone: Otter~\cite{li2023otter}. The results are shown in Tables~\ref{tab:app_otter_imagenet} and~\ref{tab:app_otter_domainbed}. We find that the performance of MVT is dependent on the MLLM backbone: when using Otter as the backbone model for MVT, the OOD performance would slightly degrade from the performance of MMICL, which could be due to the capability of MLLM to conduct ICL. However, the rectified predictions can still contain useful information to boost the performance of vision models. In several cases in ImageNet-Val, MNIST, and ImageNet-R, Otter with fine-tuning can still improve the visual robustness to the best or second-best results.

\begin{table*}[h]
	\centering
	\caption{Classification accuracy (\%) of baseline CLIP models and our method on 5 ID datasets and 5 OOD datasets. We compare the performance of our method, and the fine-tuned models supervised by our method with the baseline models, including ResNet-50 and ViT-B/32. The supervisor MLLM is MMICL~\cite{zhao2023mmicl}.}
	\setlength{\tabcolsep}{2mm}
	\begin{tabular}{l|l|ccccc|ccccc}
		\toprule
		\multirow{2}{*}{Arch}&\multirow{2}{*}{Method}&\multicolumn{5}{c|}{ID}&\multicolumn{5}{c}{OOD}\\
		\cline{3-12}
		&& IN-Val & IN-V2 & CIFAR10 & CIFAR100 & MNIST & IN-A & IN-R & IN-SK & IN-V & iWildCam \\
		\hline\hline
		\multirow{3}{*}{RN50} & CLIP & 59.7 & 52.6 & 71.5 & 41.9 & \bf58.5 & 23.9 & 60.7 & 35.4 & 31.1 & 8.2 \\
		& MVT & \bf76.2 & \bf70.8 & \bf80.2 & \bf49.7 & \underline{50.8} & \bf47.5 & \bf72.9 & \bf41.6 & \bf54.1 & \bf14.5 \\
		& +FT & \underline{66.3} & \underline{65.7} & \underline{75.1} & \underline{46.9} & 47.3 & \underline{32.1} & \underline{64.4} & \underline{36.5} & \underline{38.2} & - \\
		\hline
		\multirow{3}{*}{ViT-B} & CLIP & 62.9 & 56.1 & 89.9 & \bf65.0 & \underline{47.9} & 32.2 & 67.9 & 41.9 & 30.5 & 10.9 \\
		& MVT & \bf77.5 & \bf71.0 & \bf92.5 & \underline{60.4} & \bf51.5 & \bf60.6 & \bf83.0 & \bf47.8 & \bf53.1 & \bf19.3 \\
		& +FT & \underline{66.3} & \underline{66.0} & \underline{90.1} & 59.5 & 46.6 & \underline{38.8} & \underline{68.7} & \underline{43.1} & \underline{37.6} & - \\
		\bottomrule
	\end{tabular}
	\label{tab:app_vision_model_imagenet}
\end{table*}

\begin{table*}[h]
	\small
	\caption{Classification accuracy (\%) of baseline CLIP models and our method on 4 subsets of DomainBed datasets, including VLCS, PACS, OfficeHome, and DomainNet. We compare the performance of our method and the fine-tuned models supervised by our method with the baseline models, including ResNet-50 and ViT-B/32. The supervisor MLLM is MMICL~\cite{zhao2023mmicl}.}
	\centering
	\setlength{\tabcolsep}{1.35mm}
	\begin{tabular}{l|l|cccc|cccc|cccc|ccccc|c}
		\toprule
		\multirow{2}{*}{Arch} & Datasets & \multicolumn{4}{c|}{VLCS} & \multicolumn{4}{c|}{PACS} & \multicolumn{4}{c|}{OfficeHome} & \multicolumn{5}{c|}{DomainNet} & \multirow{2}{*}{Avg} \\
		\cline{3-19}
		& method & 0 & 1 & 2 & 3 & 0 & 1 & 2 & 3 & 0 & 1 & 2 & 3 & 0 & 1 & 2 & 3 & 4 \\
		\hline
		\hline
		\multirow{3}{*}{RN50} & CLIP & 75.0 & 82.3 & 81.3 & 75.0 & 91.3 & 90.3 & 90.0 & 96.2 & 71.7 & 80.9 & 69.4 & 67.8 & \underline{47.2} & \underline{46.8} & \underline{44.9} & \underline{64.0} & \underline{32.9} & 71.0 \\
		& MVT & \bf84.3 & \bf88.0 & \bf88.7 & \bf81.8 & \bf96.0 & \bf96.1 & \bf95.4 & \bf98.8 & \bf77.2 & \bf85.3 & \bf77.5 & \bf75.4 & \bf46.1 & \bf46.3 & \bf43.4 & \bf61.7 & \bf33.2 & \bf75.0 \\
		& +FT & \underline{83.7} & \underline{87.3} & \underline{88.1} & \underline{81.3} & \underline{95.6} & \underline{95.7} & \underline{95.1} & \underline{98.6} & \underline{75.9} & \underline{85.0} & \underline{75.8} & \underline{74.7} & 45.3 & 45.6 & 43.0 & 60.4 & 32.6 & \underline{74.3} \\
		\hline
		\multirow{3}{*}{ViT-B} & CLIP & 74.0 & 82.0 & 79.6 & 74.4 & \underline{93.6} & \underline{92.8} & \underline{93.0} & \underline{98.2} & 79.2 & 86.4 & 77.4 & 76.3 & \underline{49.7} & \underline{54.3} & \underline{51.0} & \underline{68.7} & \underline{40.7} & \underline{74.8} \\
		& MVT & \bf84.2 & \bf87.3 & \bf88.4 & \bf82.8 & \bf96.7 & \bf96.4 & \bf96.6 & \bf98.8 & \bf84.0 & \bf89.3 & \bf82.9 & \bf81.5 & \bf49.5 & \bf53.1 & \bf51.5 & \bf69.9 & \bf41.7 & \bf78.5 \\
		& +FT & \underline{76.0} & \underline{84.8} & \underline{81.3} & \underline{81.6} & 92.9 & 88.8 & 89.4 & 93.3 & \underline{81.1} & \underline{88.3} & \underline{80.8} & \underline{77.7} & 47.2 & 52.5 & 48.7 & 66.9 & 40.5 & \underline{74.8} \\
		\bottomrule
	\end{tabular}
	\label{tab:app_vision_model_domainbed}
\end{table*}

\subsection{MVT on Additional Vision Models}
\label{rn_vitb}
Then, we conduct MVT using MMICL but using different vision backbone models such as ViT-B and ResNet-50 on ImageNet and DomainBed datasets. The results are shown in Tables~\ref{tab:app_vision_model_imagenet} and~\ref{tab:app_vision_model_domainbed}. We can see that the performance of MVT is quite strong compared to other vision models which shows over $10\%$ and $4\%$ improvements in ImageNet datasets and DomainBed datasets, respectively. Especially on ImageNet-V2,  ImageNet-A, ImageNet-R, and ImageNet-V, the performance improvement of MVT are encouragingly over $15\%$, $24\%$, $12\%$, and $23\%$, respectively. After fine-tuning, the performance can be improved in most cases, such as ResNet-50 is further improved by $13.1\%$ and $3.3\%$ correspondingly on ImageNet-V2 and DomainBed thanks to MMICL.

\begin{table*}[h]
	\small
	\caption{Classification accuracy (\%) of baseline CLIP models and our method with MMICL~\cite{zhao2023mmicl} as the VLM on 15 corruptions and 5 severities of ImageNet-C datasets. We compare the performance of our method and the fine-tuned models supervised by our method with the baseline models, i.e., ViT-L from CLIP~\cite{radford2021learning}. The fine-tuned models with our MVT method have the best performance.}
	\centering
	\setlength{\tabcolsep}{1.35mm}
	\begin{tabular}{l|l|cccccc|cccccc|cccccc}
		\toprule
		\multirow{2}{*}{\shortstack{Arch}} & Datasets & \multicolumn{6}{c|}{Gaussian Noise} & \multicolumn{6}{c|}{Shot Noise} & \multicolumn{6}{c}{Impulse Noise} \\
		\cline{3-20}
		& method & 1 & 2 & 3 & 4 & 5 & avg & 1 & 2 & 3 & 4 & 5 & avg & 1 & 2 & 3 & 4 & 5 & avg \\
		\hline
		\hline
		\multirow{24}{*}{\rotatebox{90}{MMICL}} & CLIP & 69.8 & 66.7 & 59.7 & 46.9 & 30.6 & 54.7 & 70.5 & 64.9 & 57.7 & \underline{43.6} & 32.1 & 53.8 & 65.7 & 60.2 & 55.9 & 45.0 & 32.7 & 51.9 \\
		& MVT & \underline{70.1} & \underline{67.5} & \underline{61.2} & \bf49.8 & \bf{33.6} & \underline{56.4} & \underline{70.8} & \underline{66.8} & \underline{59.2} & \bf46.1 & \bf35.6 & \underline{55.7} & \underline{66.3} & \underline{61.9} & \underline{58.4} & \underline{47.5} & \underline{35.6} & \underline{53.9} \\
		& +FT & \bf71.0 & \bf67.9 & \bf61.3 & \underline{48.7} & \underline{33.5} & \bf56.5 & \bf72.0 & \bf67.1 & \bf60.1 & \bf46.1 & \underline{35.2} & \bf56.1 & \bf68.5 & \bf64.2 & \bf59.8 & \bf48.9 & \bf35.8 & \bf55.4 \\
		\cline{2-20}
		\cline{2-20}
		&& \multicolumn{6}{c|}{Defocus Blur} & \multicolumn{6}{c|}{Glass Blur} & \multicolumn{6}{c}{Motion Blur} \\
		\cline{3-20}
		&& 1 & 2 & 3 & 4 & 5 &avg& 1 & 2 & 3 & 4 & 5 &avg & 1 & 2 & 3 & 4 & 5 &avg\\
		\cline{2-20}
		\cline{2-20}
		& CLIP & 66.1 & 62.4 & 53.0 & 43.4 & 35.0 & 52.0 & 65.5 & 59.3 & 40.5 & 33.8 & 25.4 & 44.9 & 70.9 & 66.8 & 59.9 & 49.5 & 41.8 & 57.8 \\
		& MVT & \underline{67.1} & \underline{63.3} & \underline{55.8} & \bf47.6 & \bf38.8 & \underline{54.5} & \underline{67.1} & \underline{61.3} & \underline{42.8} & \underline{36.0} & \underline{29.4} & \underline{47.3} & \underline{71.9} & \underline{67.7} & \underline{60.9} & \underline{51.5} & \underline{43.2} & \underline{59.0} \\
		& +FT & \bf68.8 & \bf64.1 & \bf56.3 & \underline{47.4} & \underline{38.4} & \bf55.0 & \bf68.9 & \bf64.8 & \bf45.2 & \bf37.6 & \bf30.2 & \bf49.3 & \bf72.8 & \bf69.1 & \bf62.1 & \bf52.7 & \bf45.3 & \bf60.4 \\
		\cline{2-20}
		\cline{2-20}
		&& \multicolumn{6}{c|}{Zoom Blur} & \multicolumn{6}{c|}{Snow} & \multicolumn{6}{c}{Frost} \\
		\cline{3-20}
		&& 1 & 2 & 3 & 4 & 5 &avg& 1 & 2 & 3 & 4 & 5 &avg & 1 & 2 & 3 & 4 & 5 &avg\\
		\cline{2-20}
		\cline{2-20}
		& CLIP & 62.2 & 55.9 & 49.8 & 43.9 & 37.3 & 49.8 & 68.3 & 61.2 & 61.9 & 56.1 & 52.6 & 60.0 & 68.5 & 61.2 & 53.8 & 51.1 & 46.6 & 56.2 \\
		& MVT & \underline{64.1} & \underline{57.3} & \underline{52.0} & \underline{45.7} & \underline{38.7} & \underline{51.6} & \underline{69.2} & \underline{61.5} & \underline{62.9} & \underline{57.1} & \underline{54.0} & \underline{60.9} & \underline{69.5} & \underline{61.5} & \underline{54.2} & \underline{52.7} & \underline{47.4} & \underline{57.1} \\
		& +FT & \bf65.2 & \bf59.2 & \bf54.2 & \bf48.8 & \bf41.4 & \bf53.8 & \bf70.6 & \bf63.9 & \bf64.6 & \bf59.2 & \bf55.6 & \bf62.8 & \bf71.9 & \bf65.2 & \bf57.9 & \bf56.4 & \bf51.5 & \bf60.6 \\
		\cline{2-20}
		\cline{2-20}
		&& \multicolumn{6}{c|}{Fog} & \multicolumn{6}{c|}{Brightness} & \multicolumn{6}{c}{Contrast} \\
		\cline{3-20}
		&& 1 & 2 & 3 & 4 & 5 &avg& 1 & 2 & 3 & 4 & 5 &avg & 1 & 2 & 3 & 4 & 5 &avg \\
		\cline{2-20}
		\cline{2-20}
		& CLIP & 69.8 & 67.9 & 65.0 & 61.3 & 52.0 & 63.2 & 74.3 & 74.0 & \underline{72.8} & 70.6 & 68.1 & 72.0 & 70.6 & 69.3 & 64.8 & 52.4 & 35.1 & 58.4 \\
		& MVT & \underline{70.7} & \underline{69.2} & \underline{66.5} & \underline{62.6} & \underline{53.8} & \underline{64.6} & \underline{74.7} & \underline{74.1} & 72.6 & \underline{71.1} & \underline{68.8} & \underline{72.3} & \underline{70.9} & \underline{69.9} & \underline{65.1} & \underline{52.9} & \underline{36.9} & \underline{59.1} \\
		& +FT & \bf72.5 & \bf71.3 & \bf69.5 & \bf67.1 & \bf60.3 & \bf68.1 & \bf76.0 & \bf75.1 & \bf74.3 & \bf73.1 & \bf71.1 & \bf73.9 & \bf73.5 & \bf73.5 & \bf70.2 & \bf59.2 & \bf42.7 & \bf63.8 \\
		\cline{2-20}
		\cline{2-20}
		&& \multicolumn{6}{c|}{Elastic} & \multicolumn{6}{c|}{Pixelate} & \multicolumn{6}{c}{JPEG} \\
		\cline{3-20}
		&& 1 & 2 & 3 & 4 & 5 &avg& 1 & 2 & 3 & 4 & 5 &avg& 1 & 2 & 3 & 4 & 5 & avg \\
		\cline{2-20}
		\cline{2-20}
		& CLIP & 69.2 & 50.6 & 64.1 & 53.1 & 30.4 & 53.5 & 71.0 & 70.4 & 66.2 & 60.1 & 54.6 & 64.5 & 70.8 & 67.7 & 65.1 & 58.0 & 45.3 & 61.4 \\
		& MVT & \underline{70.0} & \underline{51.1} & \underline{65.8} & \underline{55.2} & \bf32.7 & \underline{55.0} & \underline{71.7} & \underline{70.7} & \underline{66.5} & \underline{61.9} & \underline{57.3} & \underline{65.6} & \bf72.5 & \bf69.6 & \bf67.5 & \bf60.5 & \underline{47.7} & \bf63.6 \\
		& +FT & \bf70.7 & \bf53.6 & \bf67.7 & \bf58.5 & \underline{32.2} & \bf56.5 & \bf72.8 & \bf71.8 & \bf69.1 & \bf62.9 & \bf57.7 & \bf66.9 & \underline{71.2} & \underline{68.9} & \underline{65.8} & \underline{60.0} & \bf48.8 & \underline{62.9} \\
		\bottomrule
	\end{tabular}
	\label{tab:app_corruptions}
\end{table*}

\subsection{Robustness against Visual Corruptions}
Further, we consider the visual robustness against image corruptions by evaluating our method on a robustness benchmark: ImageNet-C~\cite{hendrycks2019benchmarking}. Specifically, there are 15 different types of corruption with different corruption severities varied from 1 to 5. Here we cover all scenarios to evaluate our method using MMICL as a backbone model and a baseline method CLIP ViT-L. The results are shown in Table~\ref{tab:app_corruptions}. We can see that our method shows very strong performance in all scenarios. Compared to CLIP, using MVT can improve the performance by over $2\%$, and through fine-tuning, the performance is further boosted by over $4\%$. The encouraging results again demonstrate the effectiveness of our method.

% \subsection{**Robustness against Adversarial Attack}
\clearpage
\begin{figure}[h]
\centering
\includegraphics[height=6.3cm]{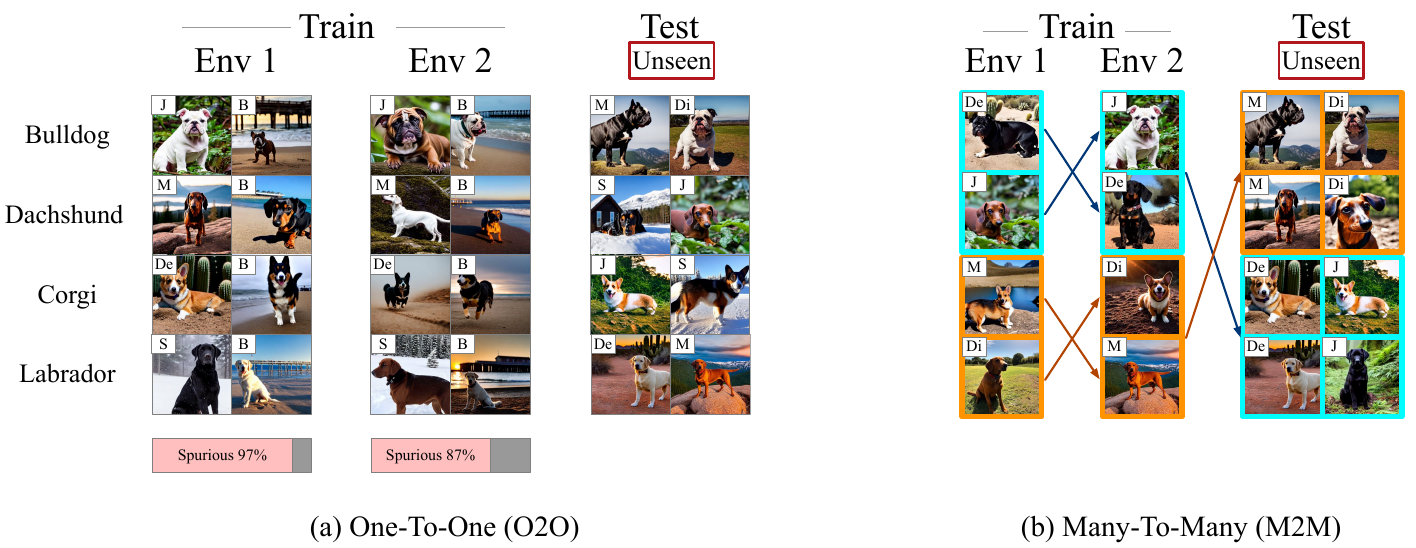}
\caption{Figures are from Lynch et al.~\cite{lynch2023spawrious}, the letters on each images denote a certain background. There are two spurious correlation types in the Spawrious dataset, namely O2O and M2M. In the O2O setting, each dog class is correlated to one certain background type and different distributions have different correlation probabilities as shown by the bar below the O2O figure. As for the M2M setting, multiple classes and backgrounds are correlated together and the correlation changes to different groups of classes and backgrounds during testing.}
\label{fig:spurious_example}
\end{figure}

\subsection{Robustness against Spurious Correlation}
Moreover, we consider a common distribution shift scenario where the training dataset and test dataset have different foreground and background correlation, \textit{i.e.}, spurious correlation. Specifically, as shown in Figure~\ref{fig:spurious_example} standing for the Spawrious dataset that we use, there are two different settings: One-To-One (O2O) correlation where each class is correlated to one background type with a certain probability. The foreground objects in the training dataset and test dataset have different probabilities to be combined with a certain background. For the Many-To-Many (M2M) setting, the foregrounds and backgrounds are split into subgroups that contain multiple classes and background types. When different subgroups are correlated together between training and test datasets, the M2M spurious correlation is formed and brings more complexity. In the Spawrious dataset, there are three levels of hardness based on correlation probability difference between training and test datasets, namely easy, medium, and hard. Here, we consider all scenarios and show the results in Table~\ref{tab:spurious}. We can see that the MVT method can outperform the ViT-L and ViT-g baseline methods in all scenarios, which leads to the conclusion that our method is robust to spurious correlations and can identify the class of interests despite the changing backgrounds.

\begin{table}[h]
	\caption{Performance comparison between MVT and CLIP on robustness against spurious correlation using Spawrious dataset.}
	\setlength{\tabcolsep}{3.5mm}
	\begin{tabular}{l|cccccc|c}
		\toprule
		Type  & O2O\_easy & O2O\_medium & O2O\_hard & M2M\_easy & M2M\_medium & M2M\_hard & Avg. \\ \hline
		ViT-L & 94.1      & 95.4        & 93.3      & 96.7      & 95.0        & 92.5      & 94.5 \\
		MVT   & \textbf{95.8}      & \textbf{96.3}        & \textbf{93.6 }     & \textbf{96.8 }     & \textbf{95.8}        & \textbf{92.9}      & \textbf{95.2} \\ \hline \hline
		ViT-g & 94.6      & 97.0        & 92.6      & 96.7      & 95.6        & 94.8      & 95.2 \\
		MVT   & \textbf{95.3 }     & \textbf{97.4}        &\textbf{ 92.8}      & \textbf{96.8}      & \textbf{96.6 }       & \textbf{95.4}      & \textbf{95.7} \\ \bottomrule
	\end{tabular}
	\label{tab:spurious}
\end{table}

\clearpage

\begin{figure}[h]
	\centering
	\includegraphics[width=\linewidth]{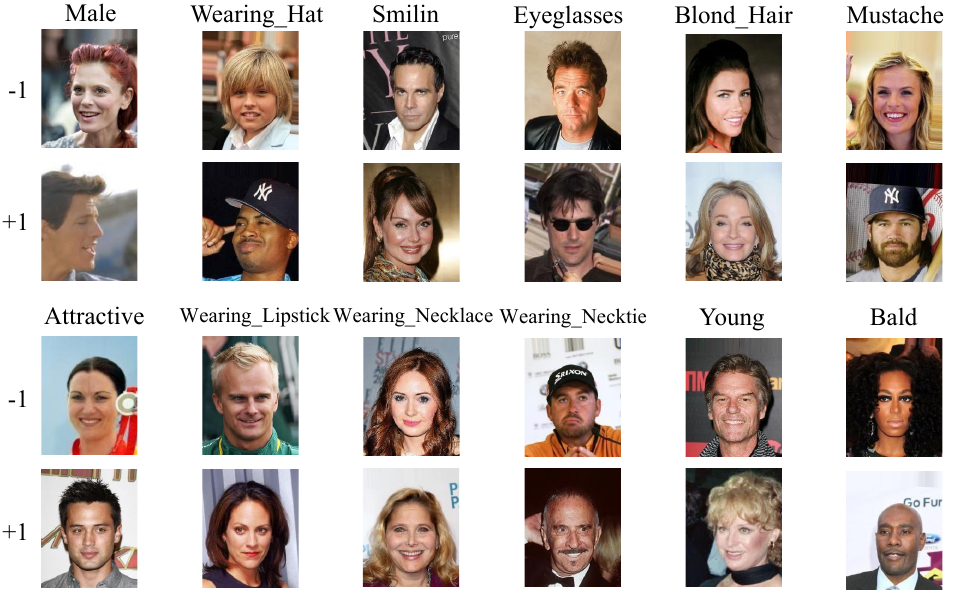}
	\caption{Examples of celebA photos with different attributes.}
	\label{fig:attribute}
\end{figure}

\begin{table}[h]
	\centering
	\caption{Class names for 12 chosen attributes.}
	\begin{tabular}{l|cc}
		\toprule
		Attribute         & \multicolumn{1}{c}{-1}                      & \multicolumn{1}{c}{+1}                  \\ \hline
		Male              & \multicolumn{1}{c}{a woman}                 & \multicolumn{1}{c}{a man}               \\
		Wear\_Hat         & \multicolumn{1}{c}{not wearing a hat}       & \multicolumn{1}{c}{wearing a hat}       \\
		Smiling           & \multicolumn{1}{c}{not smiling}             & \multicolumn{1}{c}{smiling}             \\
		Eyeglasses        & \multicolumn{1}{c}{not wearing eye glasses} & \multicolumn{1}{c}{wearing eye glasses} \\
		Blond\_Hair       & not having blond hair                       & having blind hair                       \\
		Mustache          & not having mustache                         & having mustache                         \\
		Attractive        & not attractive                              & attractive                              \\
		Wearing\_Lipstick & not wearing lipstick                        & wearing lipstick                        \\
		Wearing\_Necklace & not wearing necklace                        & wearing necklace                        \\
		Wearing\_Necktie  & not wearing necktie                         & wearing necktie                         \\
		Young             & not young                                   & young                                   \\
		Bald              & not bald                                    & bald                                    \\ \bottomrule
	\end{tabular}
	\label{tab:attribute_classnames}
\end{table}

\subsection{Performance on Recognizing Fine-grained Attributes}

Additionally, here we further explore the capability of recognizing subtle attributes based on the CelebA dataset~\cite{liu2015faceattributes}. Particularly, we consider 12 face attributes, as shown in Figure~\ref{fig:attribute}. For each attribute, we testify whether a learning model could correctly identify the attribute in a given image. Here we compare our MVT method with CLIP ViT-L and ViT-g, and the performance of MVT produced by conducting therapy on ViT-L and ViT-g models.

Particularly, since CelebA is a binary classification task, here we design different prompts for vision models and our MLLM. For CLIP models, we use \code{The person in this image is <\#classname>} as text input, where \code{<\#classname>} of each attribute is shown in Table~\ref{tab:attribute_classnames}. For our method, we still designed one positive prompt and one negative prompt for each ICL round. Specifically, for ``Male'' attribute, our in-context instruction is as follows:
\begin{tcolorbox}
	Question: Is the person in this image \{replace\_roken\} a male? Answer: True;\\
	
	Question: Is the person in this image \{replace\_roken\} a female? Answer: False;\\
	
	Question: Is the person in this image \{replace\_roken\} a male? Answer:
\end{tcolorbox}
in which is first exemplar demonstrates an image of a male positively described as male, the second exemplar shows an image of a male negatively described as female, and finally, we ask whether the input image is a male and use the output of MLLM as the prediction.

\begin{table}[h]
	\caption{Performance comparison between MVT and CLIP on recognizing fine-grained attributes using CelebA dataset.}
	\setlength{\tabcolsep}{0.85mm}
	\begin{tabular}{l|cccccccccccc|c}
		\toprule
		\scriptsize{Attr.} & \scriptsize{Male} & \scriptsize{Wearing\_Hat} & \scriptsize{Smiling} & \scriptsize{Eyeglasses} & \scriptsize{Blond\_Hair} & \scriptsize{Mustache} & \scriptsize{Attractive} & \scriptsize{Wearing\_Lipstick} & \scriptsize{Wearing\_Necklace} & \scriptsize{Wearing\_Necktie} & \scriptsize{Young} & \scriptsize{Bald} & \scriptsize{Avg.} \\ \hline
		
		ViT-L     & 63.0 & 60.8         & 64.5    & 75.8       & 36.2        & 29.0     & 42.0       & 30.8              & 38.0              & 37.5             & 66.6  & 86.3 & 52.5 \\
		MVT       & \textbf{74.0} & \textbf{67.0 }        & \textbf{65.4}    & \textbf{76.1}       & \textbf{53.0}        & \textbf{55.8}     & \textbf{42.4}       & \textbf{39.4}              & \textbf{38.6 }             & \textbf{53.9}             & \textbf{73.5}  & \textbf{88.1} & \textbf{60.6 }\\ \hline \hline
		ViT-g     & 98.5 & 75.5         & 70.4    & 83.8       & 46.0        & 66.6     & 58.2       & 72.5              & 43.5              & 28.4             & 54.1  & 91.3 & 65.7 \\
		MVT       & \textbf{98.9} & \textbf{77.2}         & \textbf{71.0}    & \textbf{84.1}       & \textbf{58.3}        & \textbf{74.9 }    & \textbf{59.0  }     & \textbf{73.2}              & \textbf{43.6}              & \textbf{41.2}             & \textbf{56.1}  & \textbf{91.8} & \textbf{69.1} \\ \bottomrule
	\end{tabular}
	\label{tab:celeba}
\end{table}

The results on CelebA are shown in Table~\ref{tab:celeba}, we observe that our method is quite effective in recognizing fine-grained attributes and its performance significantly surpasses ViT-L and ViT-g with a large margin. Especially in attributes such as  ``Blond\_Hair'', ``Mustache'', and ``Wearing\_Necktie'', the performance improvements are even over $10\%$ on both two CLIP models, and the final averaged results on all 12 attributes, the total improvements are $8.1\%$ and $3.4\%$ for ViT-L and ViT-g, respectively. Therefore, it is reasonable to conclude that our method can be effectively conducted on fine-grained attribute recognition and significantly outperforms several powerful vision models.

\section{Complementary Experimental Results}
\label{sec:complementary_exp}
Due to the non-negligible inference time of MLLMs, we cannot conduct performance evaluation of VQA and MVT on full test set of ImageNet variants and CIFAR variants. Therefore, some results in Table \ref{tab:1} are conducted under a test set split. To further fully validate our performance under the original test set, we conduct MVT on a vision model which successfully distills the knowledge and enables efficient inference. The results are shown in Table \ref{tab:full_test_set}. We can see that the effectiveness of MVT is again validated on all datasets. We can also observe a similar effect to the main paper: the performance improvement on OOD data is much more significant than that on ID data, which still validates the effectiveness of MVT on enhancing visual robustness.

\begin{table}[h]
\setlength{\tabcolsep}{2.7mm}
\caption{Performance comparison on full test set evaluation.}
\begin{tabular}{l|l|ccccccccc}
\toprule
Arch  & Method & \multicolumn{1}{c}{IN-Val} & \multicolumn{1}{c}{IN-V2} & \multicolumn{1}{c}{CIFAR10} & \multicolumn{1}{c}{CIFAR100} & \multicolumn{1}{c}{MNIST} & IN-A & IN-R & IN-SK & IN-V \\ \hline
\multirow{2}{*}{ViT-L} & CLIP   & 75.5                       & 69.8                      & 95.5                        & 78.3                         & 76.4                      & 70.8 & 87.8 & 59.6  & 51.5 \\
 & MVT+FT & 77.1                       & 70.4                      & 98.1                        & 81.8                         & 79.3                      & 75.3 & 89.9 & 61.9  & 68.6 \\ \hline
\multirow{2}{*}{ViT-g} & EVA    & 80.1                       & 73.6                      & 98.3                        & 88.7                         & 62.5                      & 69.4 & 92.2 & 68.9  & 64.9 \\
 & MVT+FT & 81.2                       & 74.9                      & 98.6                        & 90.9                         & 65.6                      & 75.0 & 93.9 & 71.1  & 70.6 \\ \bottomrule
\end{tabular}
\label{tab:full_test_set}
\end{table}

\clearpage
\section{Additional Performance Analysis}
\label{sec:app_analysis}
In this section, we carefully conduct additional performance analysis to further validate the effectiveness of our MVT.

\subsection{Analysis on MLLM Guided Fine-Tuning}
\begin{figure}[H]
\vspace{-3mm}
\centering
\includegraphics[width=0.85\linewidth]{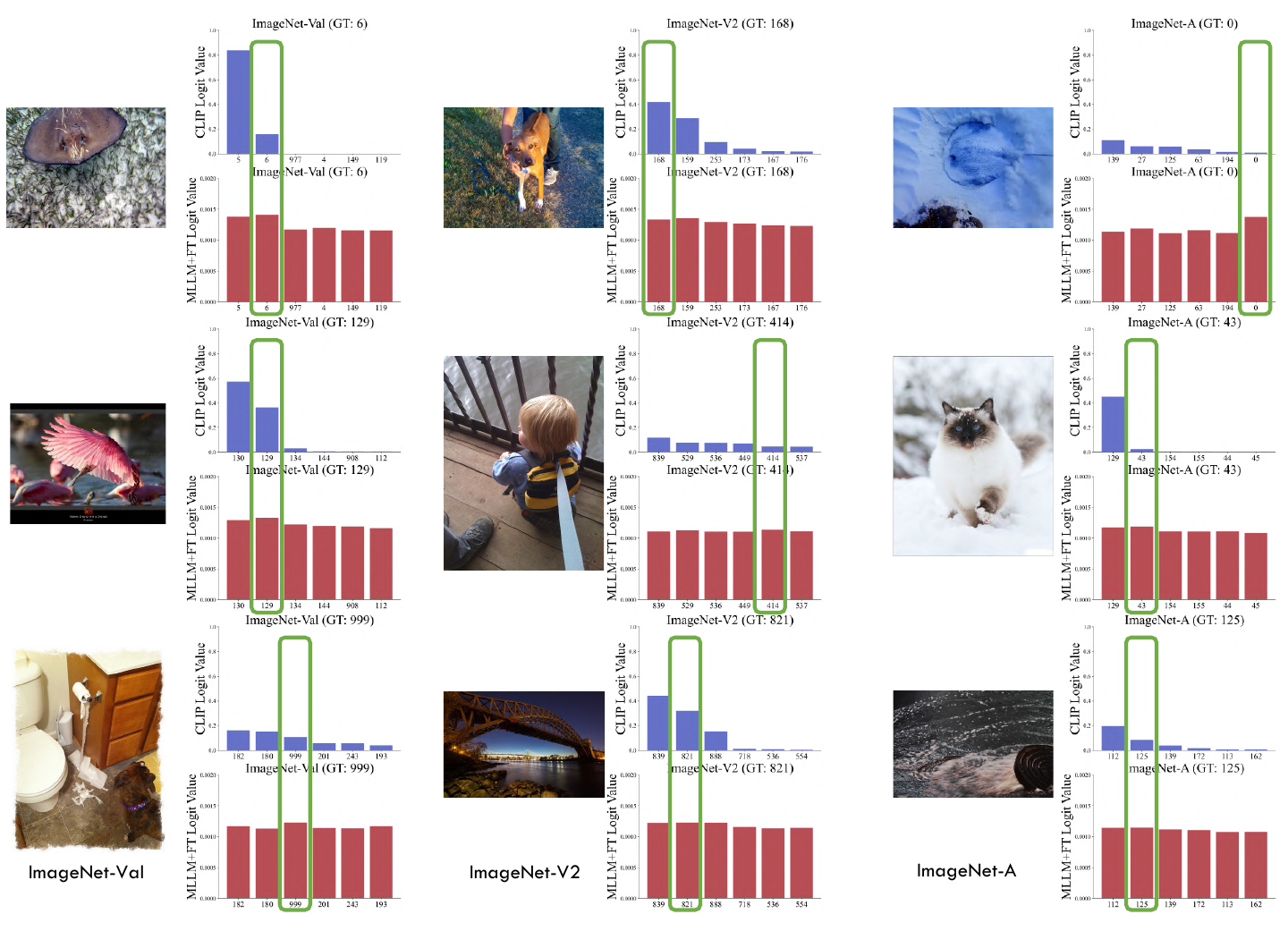}
\centering
\includegraphics[width=0.85\linewidth]{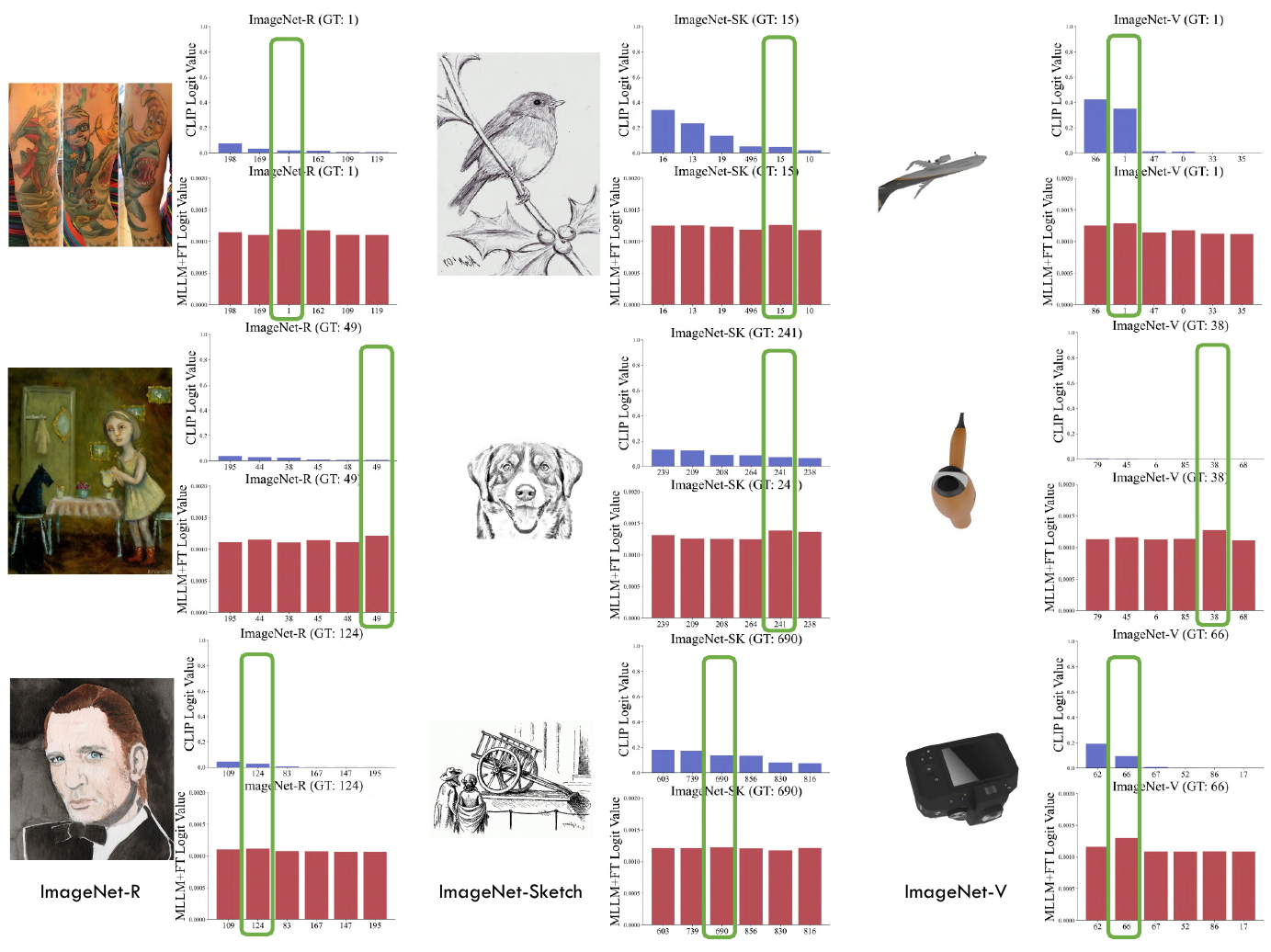}
\caption{Images with the prediction logits before and after fine-tuning. Examples are randomly chosen from ImageNet-Val, ImageNet-V2, ImageNet-A, ImageNet-R, ImageNet-Sketch, and ImageNet-V. The green boxes highlight the ground truth class logits.}
\label{fig:fine_tune_1_2}
\end{figure}

We find that our MLLM-guided fine-tuning is quite effective in further improving the prediction accuracy based on MVT corrections. To investigate why such a fine-tuning process can help the learning performance, here we compare the prediction logits of image examples from ImageNet and its variant datasets before and after the fine-tuning process. As we already shown in Figure 4 in the main paper, our MVT framework can effectively find the examples misclassified by the vision model, thus we randomly select some images that are further processed with our therapy and fine-tuning. The logits are shown in Figure~\ref{fig:fine_tune_1_2}, we can observe that after fine-tuning, the predictions of the previously incorrect examples are finally rectified which shows that the proposed MVT method is indeed helpful for visual recognition. Moreover, we also observe that the ground truth logit values of some examples are quite small which means that the vision models are very confident in their incorrect predictions. Thanks to our MVT, we can not only successfully find the ground truth classes, but also help produce a less confident prediction for the previously incorrect examples. We hypothesize such an effect could help mitigate the overfitting issue~\cite{lin2023over, lin2024eliminating} and thus generalize better on unseen OOD test sets.

\subsection{Analysis on Various DICL Designs}
To justify why using just one positive-negative exemplar pair can effectively conduct vision tasks, here we provide an analysis of using three different DICL designs. Specifically, we consider using two positive exemplars, two negative exemplars, and two incorrect exemplars as baseline DICL designs and compare them to the original MVT results in the main paper. For example, the three types of instructions are shown as:

Two positive exemplars:
\begin{tcolorbox}
	Question: This image $<$IMG\_PRE\#0$>$ shows a photo of $<$PRE\#0$>$, True or False? Answer: True;\\
	\vspace{-2.5mm}
	
	Question: This image $<$IMG\_CLN\#c$>$ shows a photo of $<$CLN\#c$>$, True or False? Answer: True;\\
	\vspace{-2.5mm}
	
	Question: This image $<$IMG\_Query$>$ shows a photo of $<$PRE\#0$>$, True or False? Answer:
\end{tcolorbox}
Two negative exemplars:
\begin{tcolorbox}
	Question: This image $<$IMG\_CLN\#c$>$ shows a photo of $<$PRE\#0$>$, True or False? Answer: False;\\
	\vspace{-2.5mm}
	
	Question: This image $<$IMG\_CLN\#c+1$>$ shows a photo of $<$PRE\#0$>$, True or False? Answer: False;\\
	\vspace{-2.5mm}
	
	Question: This image $<$IMG\_Query$>$ shows a photo of $<$PRE\#0$>$, True or False? Answer:
\end{tcolorbox}
Two incorrect exemplars:
\begin{tcolorbox}
	Question: This image $<$IMG\_CLN\#c$>$ shows a photo of $<$PRE\#0$>$, True or False? Answer: True;\\
	\vspace{-2.5mm}
	
	Question: This image $<$IMG\_PRE\#0$>$ shows a photo of $<$PRE\#0$>$, True or False? Answer: False;\\
	\vspace{-2.5mm}
	
	Question: This image $<$IMG\_Query$>$ shows a photo of $<$PRE\#0$>$, True or False? Answer:
\end{tcolorbox}

\begin{table*}[h]
	\vspace{-2mm}
	\centering
	\caption{Performance of using different DICL prompt designs. We also compare them with the original MVT using ViT-L and fine-tuning performance using the MMICL backbone.}
	\setlength{\tabcolsep}{1.5mm}
	\begin{tabular}{c|l|ccccc|ccccc}
		\toprule
		\multirow{2}{*}{MLLM}&\multirow{2}{*}{Method}&\multicolumn{5}{c|}{ID}&\multicolumn{5}{c}{OOD}\\
		\cline{3-12}
		&& IN-Val & IN-V2 & CIFAR10 & CIFAR100 & MNIST & IN-A & IN-R & IN-SK & IN-V & iWildCam \\
		\hline\hline
		None & CLIP & \underline{75.8} & 70.2& 95.6 & 78.2 & \underline{76.4} & 69.3 & 86.6 & \underline{59.4} & 51.8 & 13.4 \\
		\hline
		
		\multirow{5}{*}{\rotatebox{90}{MMICL}} 
		& Two incorrect & 73.2 & 68.3& 93.2 & 77.6 & 54.6 & 68.7 & 86.8& 58.2 & 58.3 & 22.3 \\
		& Two positive & 74.5 & 69.5& 94.1 & 77.9 & 56.2 & 69.0 & 87.4 & 57.5 & 60.9 & 23.7 \\
		& Two negative & 75.0 & 69.0& 96.6 & 78.1 & 55.7 & 69.3 & 87.8 & 58.1 & 61.2 & \underline{24.3} \\
		
		\cline{2-12}
		
		& MVT & 75.2 & \bf70.8 & \bf97.9 & \underline{78.9} & 53.0 & \underline{71.2} & \underline{88.1} & 59.0 & \underline{62.1} & \bf25.0 \\
		& +FT & \bf76.9 & \underline{70.5} & \underline{96.7} & \underline{82.0} & \bf79.2 & \bf75.1 & \bf89.5 & \bf61.4 & \bf68.8 & - \\
		\bottomrule
	\end{tabular}
	\label{tab:dicl_design}
\end{table*}

The results are shown in Table~\ref{tab:dicl_design}. We find that all three designs are inferior to our method MVT, thus we know that using one positive and negative exemplar pair is the most effective instruction strategy. Moreover, we find that using Two negative exemplars is slightly better than the other two, which manifests that negative exemplars are quite important in deciding the effectiveness of MVT, further justifying that using a noise transition matrix to find the most probable negative classes is essential to our method. Furthermore, we also find that using two incorrect exemplars achieves the worst performance, even inferior to CLIP ViT-L. This could be because that feed incorrect instructions could indeed mislead the learning performance, thus showing degradation.

\subsection{Analysis on OOD Robustness}
\begin{figure}[H]
\vspace{-0mm}
\centering
\includegraphics[width=0.98\linewidth]{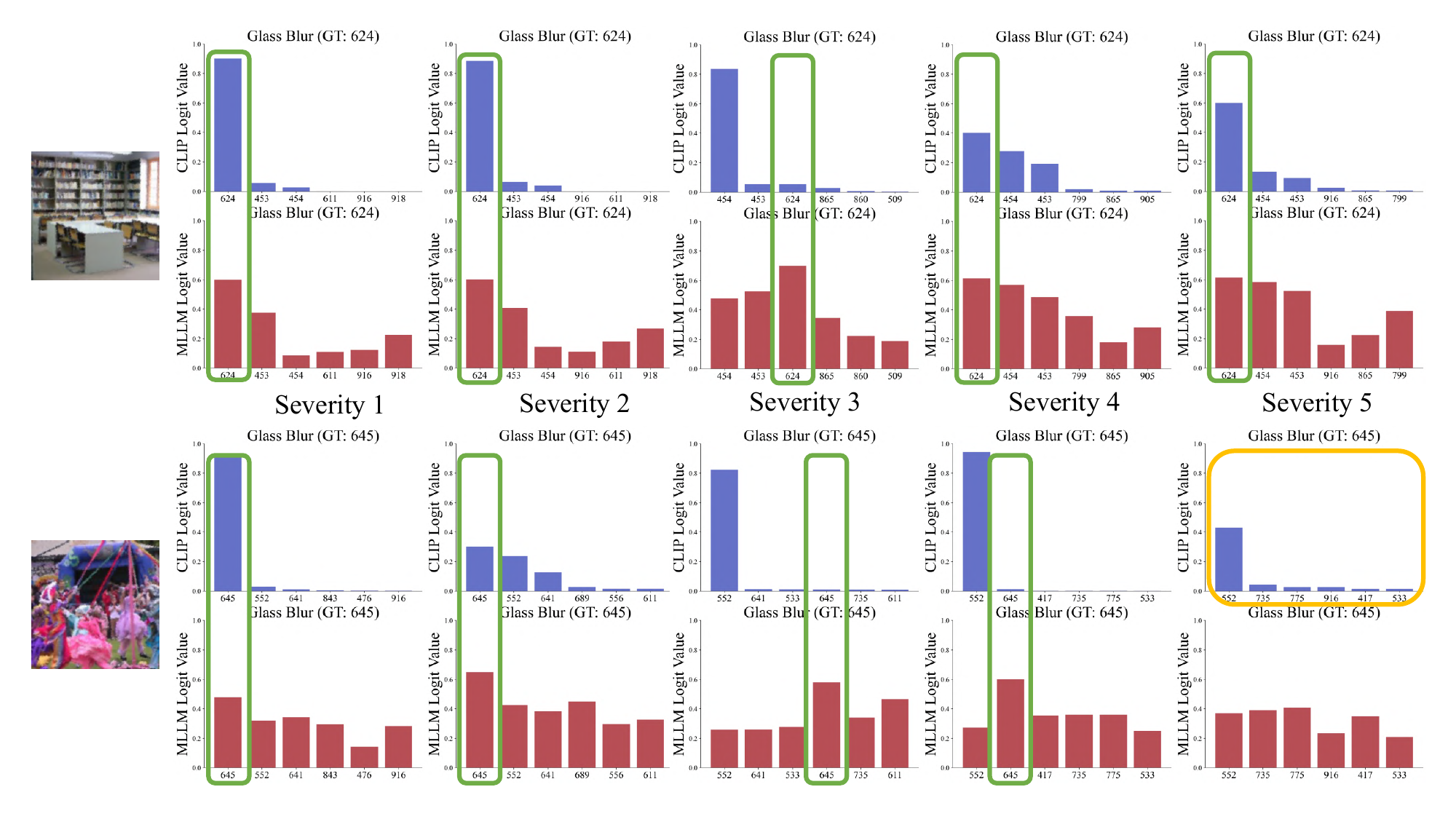}
\centering
\includegraphics[width=0.98\linewidth]{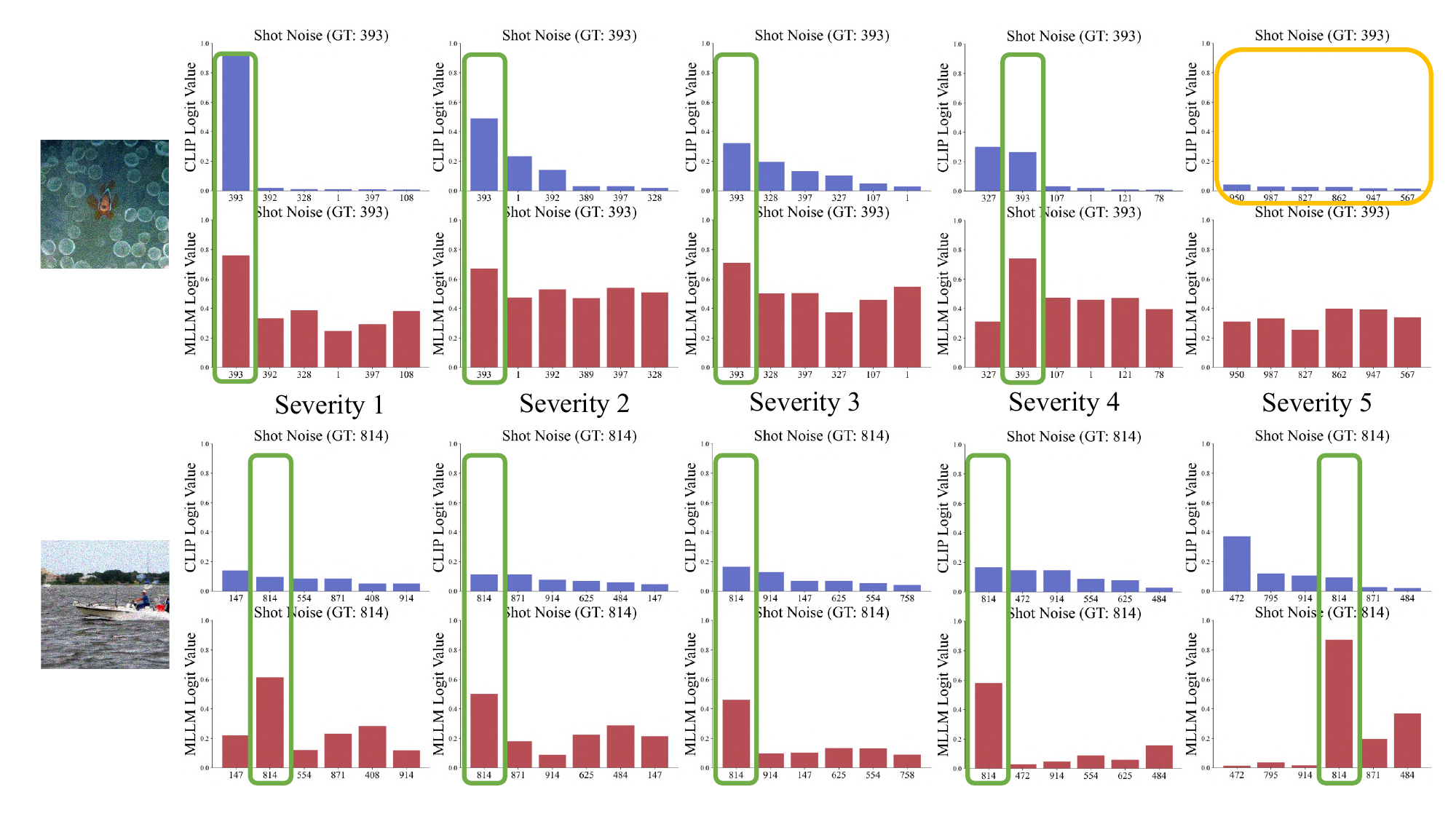}
\caption{Robustness analysis using ImageNet-C. The first two rows are examples of Glass Blur corruption, and the last two rows are examples of Shot Noise corruption. The green boxes highlight the logits of ground truth classes, and the orange boxes denote there are no ground truth predictions in the CLIP top-$6$ predictions.}
\label{fig:ood_robustness_1_2}
\end{figure}

To further investigate the performance on facing OOD data with varied strengths, here we use ImageNet-C to show how increasing the corruption severity could affect the prediction of the vision model and MLLM (MMICL). For illustration, we randomly choose four examples from ImageNet-C and plot their top-$6$ prediction logits from the vision model. Moreover, for clear comparison, we also choose the top-$6$ prediction classes as our therapy candidates (which is different from the settings of MVT) to show the MLLM prediction results. As shown in Figure~\ref{fig:ood_robustness_1_2}, we can see that as the severity increases, the prediction of CLIP logit values is highly unstable. When the severity is large, the final top-$1$ prediction could be incorrect. However, the prediction of MLLM remains consistent through all severities. Even when CLIP prediction is incorrect, MLLM can still correctly find the ground truth classes. Only when there is no ground truth in top-$6$ predictions, MLLM can be uncertain about the final prediction. Therefore, the robustness of MLLM against corruption could justify that our MVT is quite effective on OOD tasks compared to vision models.

\begin{figure}[H]
\centering
\includegraphics[width=\linewidth]{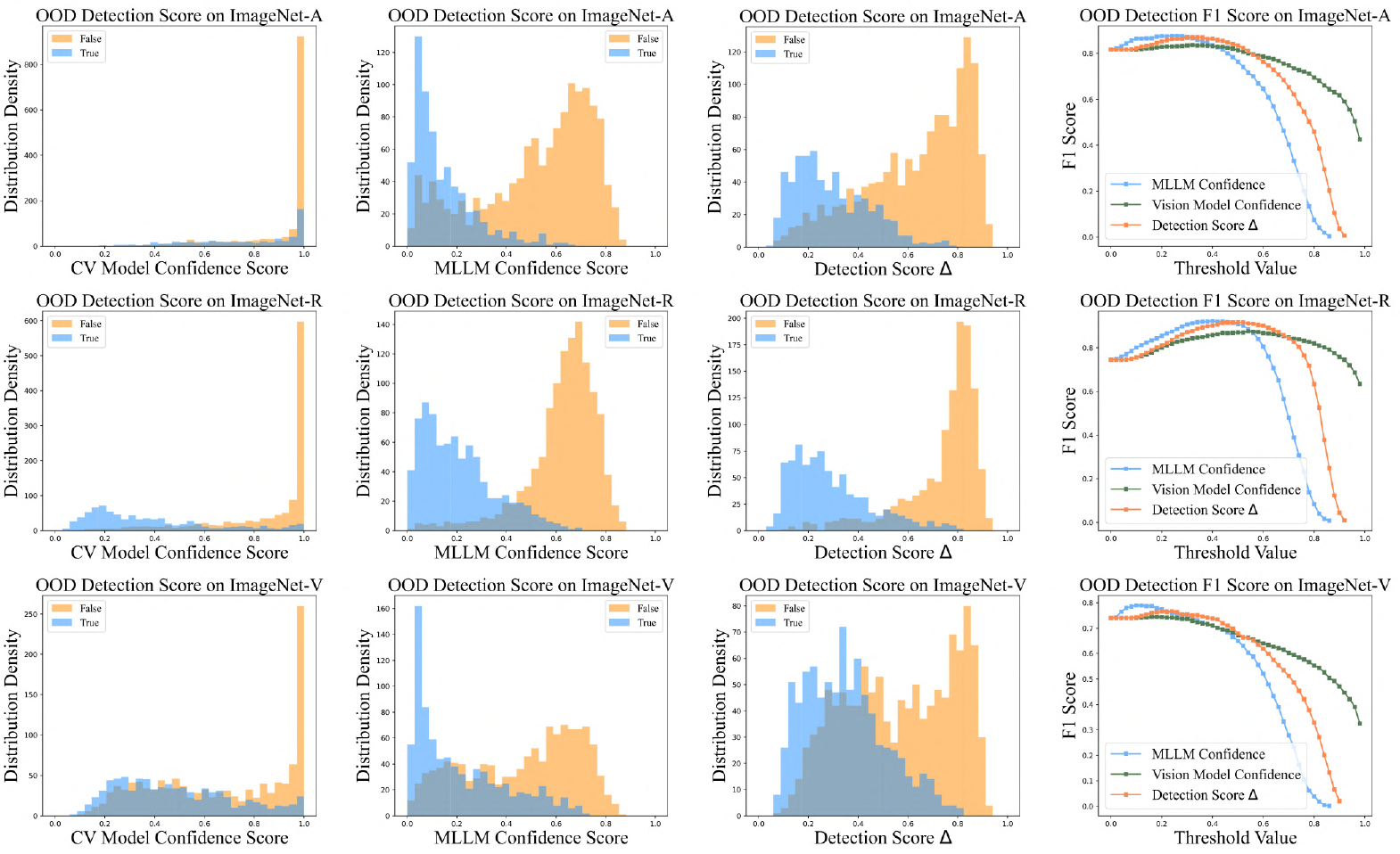}
\centering
\includegraphics[width=\linewidth]{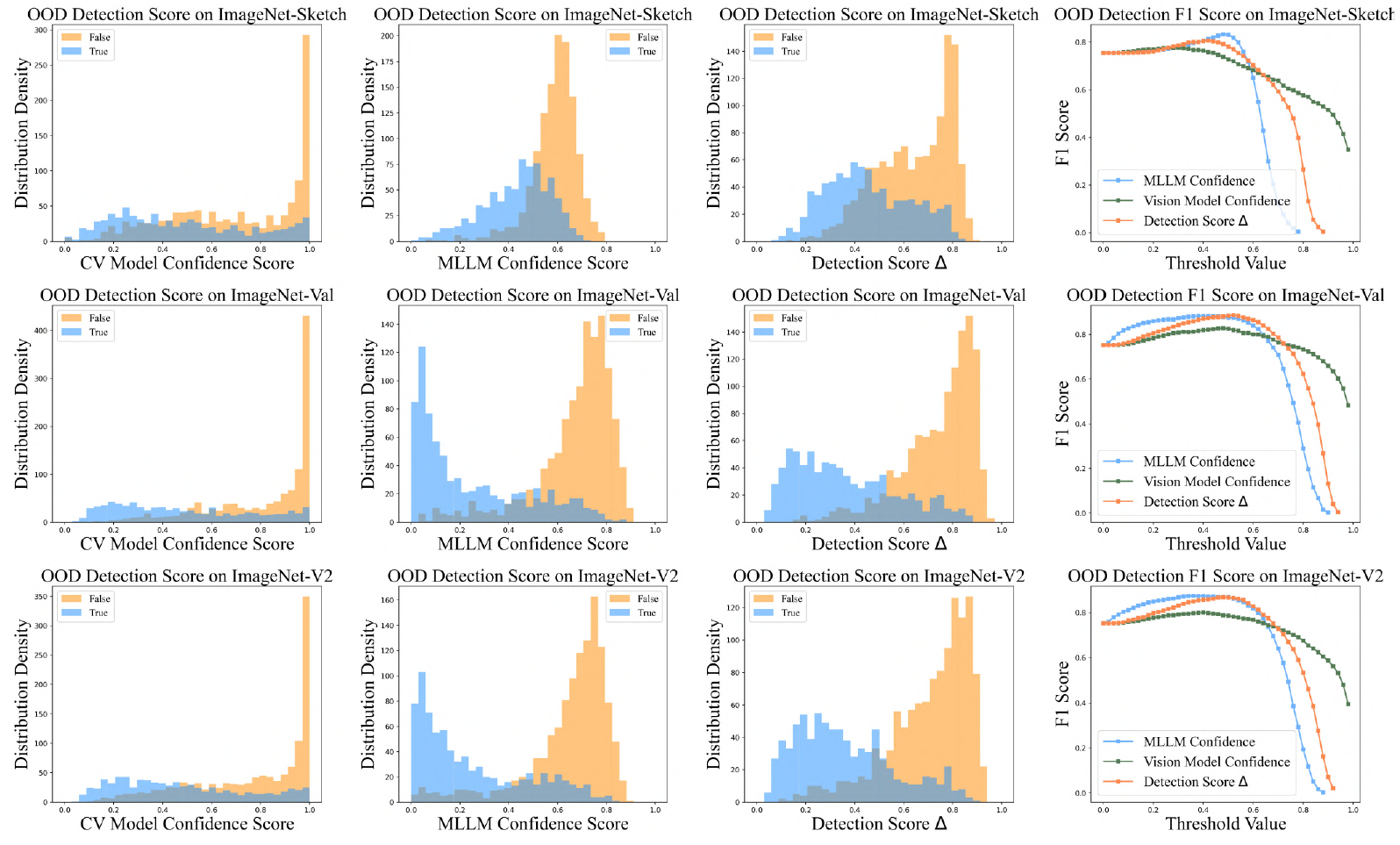}
\caption{OOD detection performance on ImageNet-A, ImageNet-R, ImageNet-V, ImageNet-Sketch, ImageNet-Val, and ImageNet-V2 datasets using ViT-L. The first three columns: Density distribution of different OOD detection scores; The last column: F1 score values by varying the OOD detection threshold $\delta$.}
\label{fig:ood_detection_1_2}
\end{figure}

\begin{figure}[H]
	\centering
	\includegraphics[width=\linewidth]{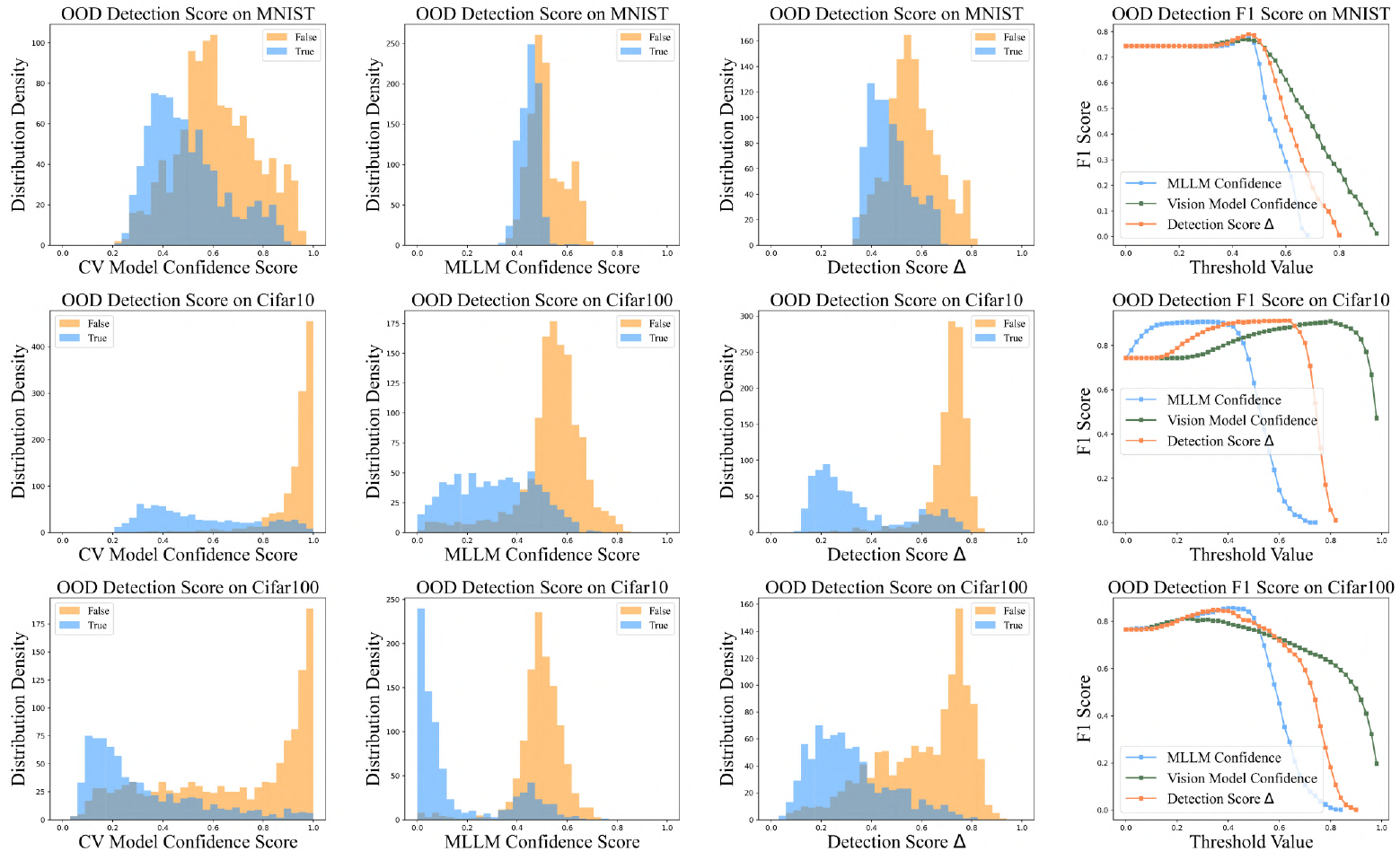}
	\centering
	\includegraphics[width=\linewidth]{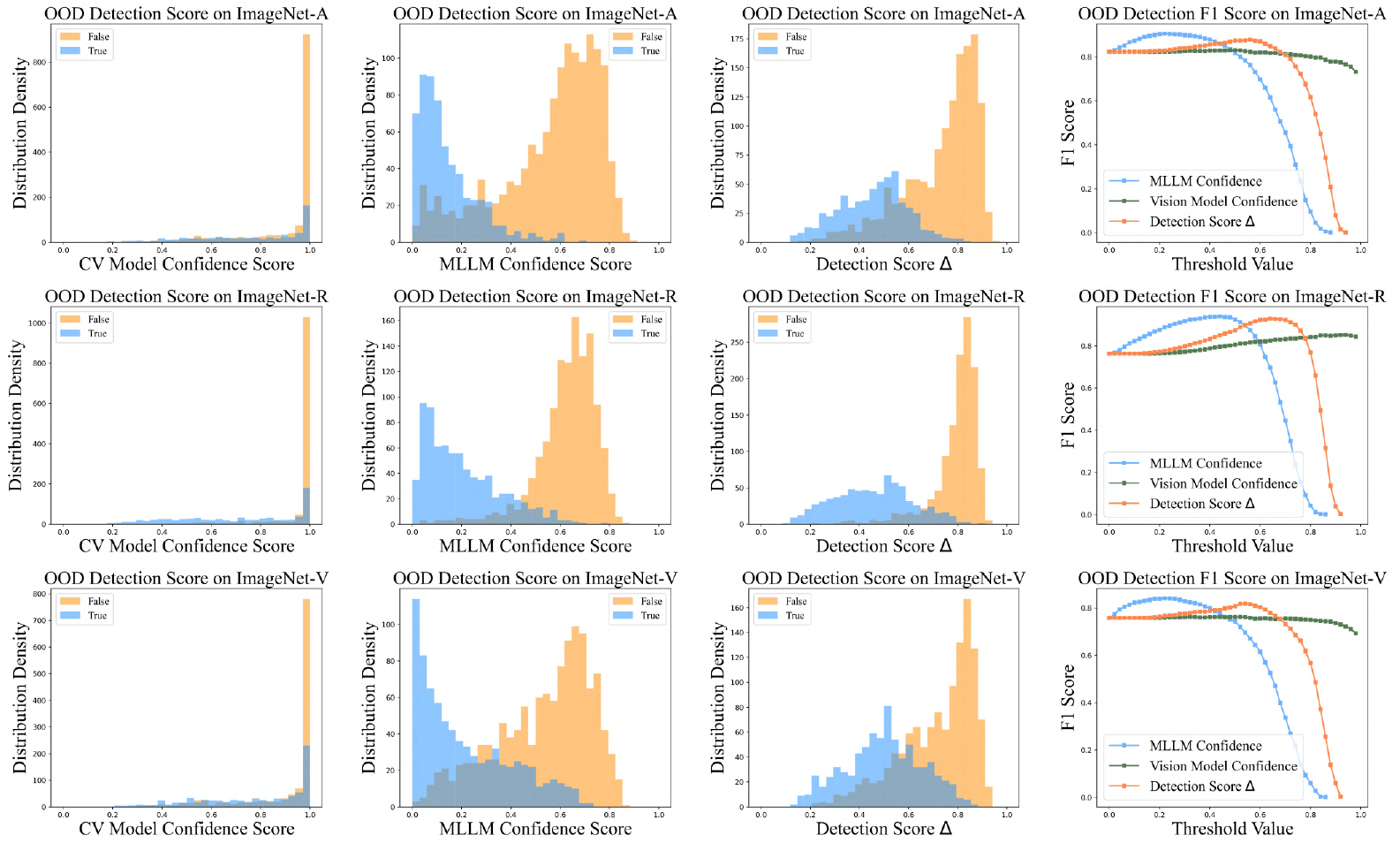}
	\caption{First three rows: OOD detection performance on MNIST, CIFAR10, and CIFAR100 datasets using ViT-L. Last three rows: OOD detection performance on ImageNet-A, ImageNet-R, and ImageNet-V datasets using ViT-g. The first three columns: Density distribution of different OOD detection scores; The last column: F1 score values by varying the OOD detection threshold $\delta$.}
	\label{fig:ood_detection_3_vitg_1}
\end{figure}

\begin{figure}[H]
	\centering
	\includegraphics[width=\linewidth]{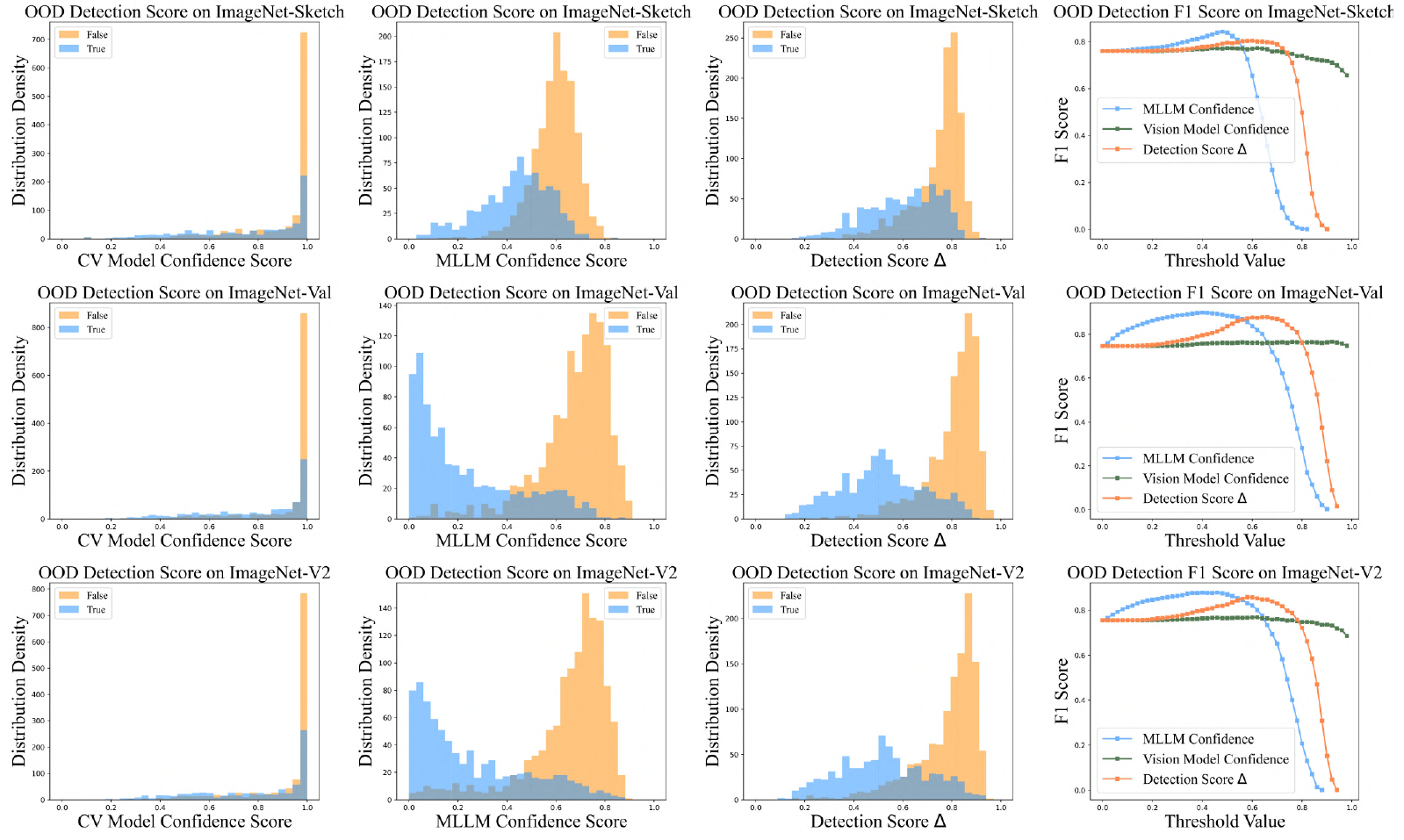}
	\centering
	\includegraphics[width=\linewidth]{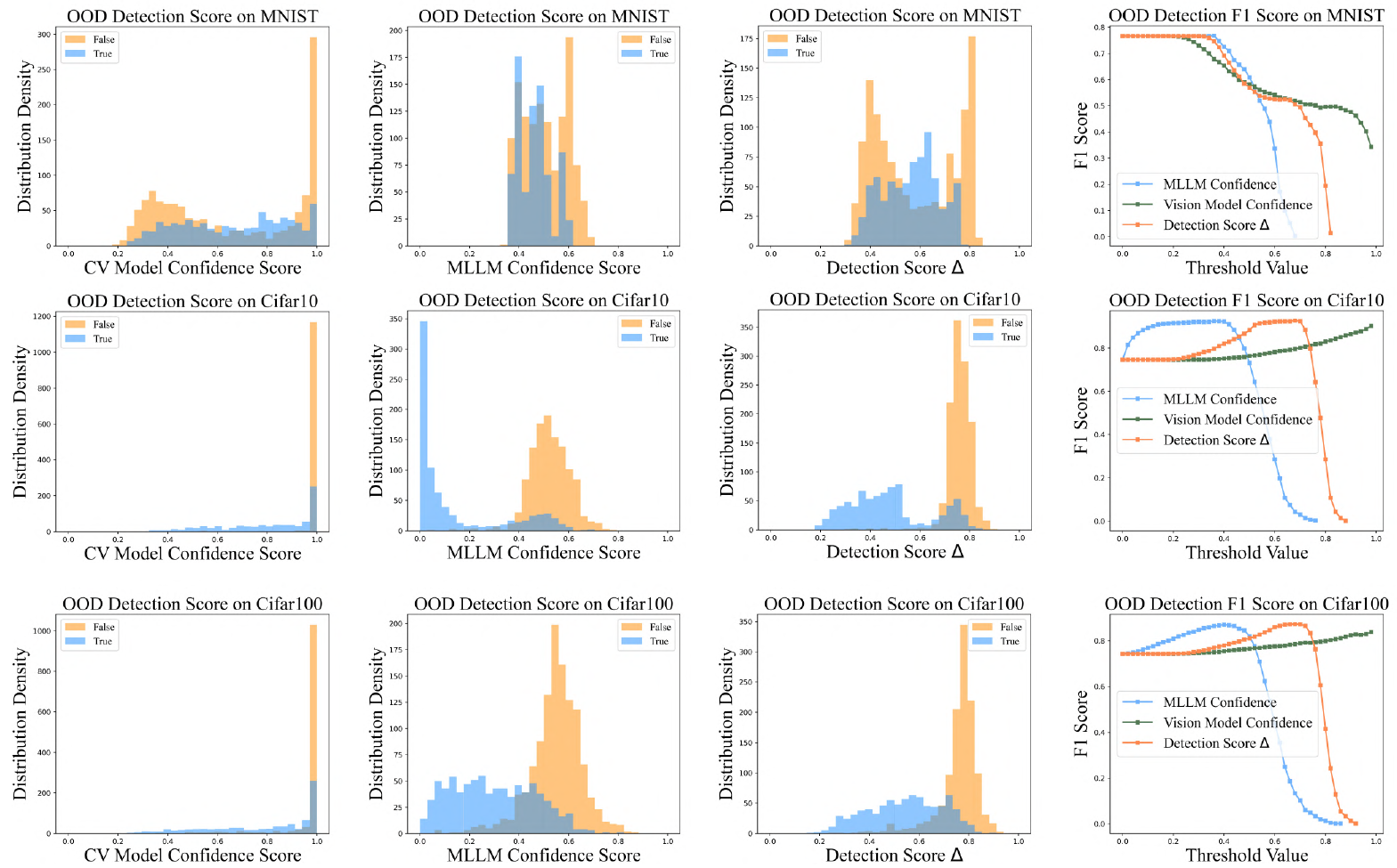}
	\caption{OOD detection performance on ImageNet-Sketch, ImageNet-Val, ImageNet-V2, MNIST, CIFAR10, and CIFAR100 datasets using ViT-g. The first three columns: Density distribution of different OOD detection scores; The last column: F1 score values by varying the OOD detection threshold $\delta$.}
	\label{fig:vitg_ood_detection_2_3}
\end{figure}

\subsection{Analysis on OOD Detection}
As an extension of the main paper, here we further provide more results on OOD detection. We consider ImageNet-A, ImageNet-R, ImageNet-V, ImageNet-Sketch, ImageNet-Val, ImageNet-V2, MNIST, CIFAR10, and CIFAR100 datasets using both ViT-L and ViT-g models. The results are shown in Figures~\ref{fig:ood_detection_1_2},~\ref{fig:ood_detection_3_vitg_1},~\ref{fig:vitg_ood_detection_2_3}. We can observe similar phenomenons as in the main paper: MLLM can effectively identify open-class data and vision models can identify close-class data. By combining the prediction confidence of MLLM and vision models, our detection score $\Delta$ can be successfully leveraged to conduct OOD detection, and the F1 score of $\Delta$ can achieve the largest value with a reasonable detection threshold $\delta$. Still, some datasets are relatively challenging compared to other datasets: ImageNet-Sketch and MNIST. This could be due to that the classification performance on these two datasets is not outstanding, about $50\%$ to $68\%$ for both vision models and MLLMs. Moreover, the patterns of such two datasets are quite simple: both of them are handwritten lines without complex natural features, which could further hinder the extraction of useful knowledge, thus leading to sub-optimal detection performance. Overall, the OOD detection performance of MVT is effective on most datasets and the capability of recognizing unknown examples could provide insight to the OOD detection field. We plan to further investigate its potential in future works.

\section{Limitation and Broader Impact}
\label{sec:app_limitation}
In this paper, we proposed an effective framework that aims to enhance the visual robustness of vision models by exploiting the knowledge of MLLMs instead of requiring additional human annotations. Based on our proposed DICL strategy, the paradigm of MLLMs can be perfectly aligned to vision tasks and achieve encouraging results. However, we found that the performance of our MVT is highly related to the ICL capability of MLLMs. Moreover, since there are only two existing MLLMs that possess multimodal ICL power, the potential of our MVT framework could be further improved when sophisticated MLLMs with multimodal ICL abilities are developed in the future. We hope our work could bring insight into the multimodal learning field with our DICL to achieve alignment between MLLMs and vision learning tasks. Based on our work, we believe many traditional fields that are related to visual recognition such as weakly-supervised learning, OOD detection, and fine-grained image classification could be further advanced by effectively leveraging the knowledge of MLLMs.

\clearpage